\documentclass[10pt,journal,compsoc]{IEEEtran}
\usepackage{multirow}
\usepackage{amssymb}
\usepackage{graphicx}
\usepackage{color}
\usepackage{url}  
\ifCLASSOPTIONcompsoc
  \usepackage[nocompress]{cite}
\else
  \usepackage{cite}
\fi

\usepackage{tablefootnote}
\usepackage{footnotehyper}
\usepackage[numbers,sort&compress]{natbib}

\definecolor{airforceblue}{rgb}{0.36, 0.54, 0.66}

\newcommand{\zliu}[1]{\textcolor{black}{#1}}
\newcommand{\zliuall}{\color{black}}
\newcommand{\zliublack}[1]{\textcolor{black}{#1}}
\newcommand{\zliunew}[1]{\textcolor{black}{#1}}

\begin{document}
%

\title{EPNet++: Cascade Bi-directional Fusion for  Multi-Modal 3D Object Detection}

\author{Zhe~Liu,
        Tengteng~Huang,
        Bingling~Li,
        Xiwu~Chen,
        Xi~Wang,
        Xiang~Bai
\IEEEcompsocitemizethanks{
\IEEEcompsocthanksitem Z. Liu, B. Li and X. Chen are with the School of Electronic Information and Communications, Huazhong University of Science and Technology, Wuhan, 430074, China. \protect \\
E-mail: \{zheliu1994, blli, xiwuchen\}@hust.edu.cn
\IEEEcompsocthanksitem T. Huang is with Megvii (Face++) Inc., Beijing, 100190, China. \protect \\
E-mail: tengtenghuang@foxmail.com.
\IEEEcompsocthanksitem X. Wang is the founder and CEO of CalmCar, Suzhou, 215168, China. \protect \\
E-mail: xi.wang@calmcar.com.
\IEEEcompsocthanksitem X. Bai is with the School of Artificial Intelligence and Automation, Huazhong University of Science and Technology, Wuhan, 430074, China. \protect \\
E-mail: xbai@hust.edu.cn.}}


\IEEEtitleabstractindextext{%
\begin{abstract}
Recently, fusing the LiDAR point cloud and camera image to improve the performance and robustness of 3D object detection has received more and more attention, as these two modalities naturally possess strong complementarity. 
In this paper, we propose EPNet++ for multi-modal 3D object detection by introducing a novel Cascade Bi-directional Fusion~(CB-Fusion) module and a Multi-Modal Consistency~(MC) loss. 
\zliu{More concretely, the proposed CB-Fusion module enhances point features with plentiful semantic information absorbed from the image features in a cascade bi-directional interaction
fusion manner, leading to more powerful and discriminative feature representations.}
The MC loss explicitly guarantees the consistency between predicted scores from two modalities to obtain more comprehensive and reliable confidence scores. The experimental results on the KITTI, JRDB and SUN-RGBD datasets demonstrate the superiority of EPNet++ over the state-of-the-art methods. 
Besides, we emphasize a critical but easily overlooked problem, which is to explore the performance and robustness of a 3D detector in a sparser scene.
Extensive experiments present that EPNet++ outperforms the existing SOTA methods with remarkable margins in highly sparse point cloud cases, which might be an available direction to reduce the expensive cost of LiDAR sensors. Code is available at: \url{https://github.com/happinesslz/EPNetV2}.
\end{abstract}

}

\maketitle

\IEEEdisplaynontitleabstractindextext

%
\IEEEpeerreviewmaketitle

\IEEEraisesectionheading{\section{Introduction}\label{sec:introduction}}
\IEEEPARstart{3}D \zliu{object detection serves as one fundamental technique in self-driving, whose main research can be classified into three categories based on the input modalities: image~\cite{ma2019accurate,qin2019monogrnet,ku2019monopsr,li2019gs3d,liu2019deep,liu2021autoshape, licvpr2019, wangcvpr2019,chen2016monocular,shi2021geometry,mousavian20173d}, LiDAR~\cite{chen2017multi,zhou2018voxelnet,yang2018pixor,shi2019pointrcnn,lang2019pointpillars,yin2021center,mao2021voxel,zheng2021se,lasernet,guan2022m3detr} and multi-sensor fusion~\cite{sindagi2019mvx,yoo20203d,qi2020imvotenet,liang2018deep,vora2020pointpainting,huang2020epnet,xie2020pi,xu2018pointfusion,wang2021pointaugmenting,piergiovanni20214d,xu2017pointfusion,zhao20193d,liang2019multi}.}
In general, LiDAR-based methods\zliu{~\cite{chen2017multi,zhou2018voxelnet,yang2018pixor,shi2019pointrcnn,lang2019pointpillars,yin2021center,mao2021voxel,zheng2021se,lasernet,guan2022m3detr}} usually achieve superior 3D object detection performance over image-based\zliu{~\cite{ma2019accurate,qin2019monogrnet,ku2019monopsr,li2019gs3d,liu2019deep,liu2021autoshape, licvpr2019, wangcvpr2019,chen2016monocular,shi2021geometry,mousavian20173d}}, as image-based methods can not obtain the accurate localization of 3D objects due to the lack of depth and 3D structural information. However, LiDAR-based methods often produce false-positive detection results, especially when the geometric structure of a background object is similar to the 3D objects to be detected. Moreover, LiDAR-based methods may fail to detect distant or small objects, where the point clouds are too sparse even if collected by a high-quality LiDAR. Yet, distant or small objects are easier to be detected with the help of appearance cues in 2D camera images.  Motivated by these observations, our intuition is to design an effective deep learning framework for fully exploiting the complementary semantics of camera images and point clouds. However, integrating the cues of camera images and point clouds for 3D object detection is non-trivial due to their quite different data representations.

\begin{figure}[t]
\centering
  \includegraphics[width=0.96\linewidth]{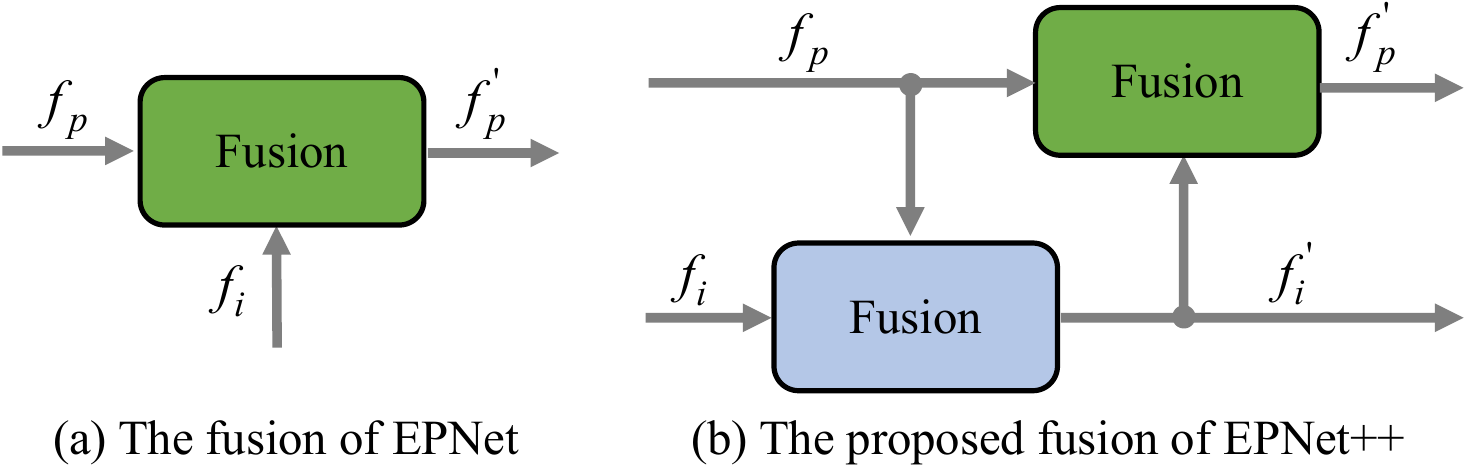}
\caption{Comparison of the fusion approach in EPNet and that in this paper. $f_i$ and $f_p$ represent the image feature and point feature, while $f_i'$ and $f_p'$ denote the enhanced image feature and enhanced point feature.}
\label{fig:intro}
\end{figure}




The previous works on multi-sensor fusion follow the pipeline of sequential fusion~\cite{xu2017pointfusion,zhao20193d,vora2020pointpainting} or multi-sensor input reasoning~\cite{liang2019multi,liang2018deep,yoo20203d}.  \zliu{The former approaches train the networks with images and point clouds separately and implement the multi-modal fusion by connecting the two stages. The final detection performance is highly dependent on the performance of each stage, posing extra optimization difficulty for the fusion module to coordinate both modalities.}
The latter methods~\cite{liang2019multi,liang2018deep,yoo20203d} alternatively study the fusion of camera images and Bird's-Eye View~(BEV) images, which are generated from point clouds through the operations of perspective projection and voxelization. \zliu{However, the 3D information of point clouds on the height dimension is usually lost when performing BEV projection. Besides, a voxel feature usually comes from the aggregation of multiple point features. Therefore, the correspondence between voxels and image features is coarser than that based on the raw point cloud. These two disadvantages might limit their upper bound on detection tasks.} 




\zliu{Different from previous methods~\cite{xu2017pointfusion,zhao20193d,vora2020pointpainting,liang2019multi,liang2018deep,yoo20203d}, our preliminary work named EPNet~\cite{huang2020epnet} proposed a \emph{LiDAR-guided Image Fusion}~(LI-Fusion) module to enrich the semantic information of point clouds for improving 3D detection. The main characteristic is that EPNet fuses the deep features from raw point clouds and camera images at different levels of resolution in a fine-grained point-wise paradigm, which can be trained in an end-to-end manner.} Though effective, EPNet simply utilizes image features for enhancing point features in a point-wise manner as shown in Fig.~\ref{fig:intro}~(a), which may ignore the benefits of the bi-directional feature interaction of both modalities. \zliu{Intuitively, point features can provide important geometric cues~(\textit{e.g.}, the contour and depth of objects) for image features, which can assist the image network to achieve more accurate object segmentation and in turn further promote the point cloud network.}


In this paper, we investigate a bi-directional fusion mechanism for multi-sensor 3D object detection, \zliu{which is named \emph{Cascade Bi-directional Fusion}~(CB-Fusion) module as illustrated in Fig.~\ref{fig:intro}~(b). We adopt point features to enhance image features first and then reinforce point features. Although the manner of feature enhancement in a reversed order is also an option, we select the order in Fig.~\ref{fig:intro}~(b). Since we use point features instead of image features to predict 3D boxes, it is more reasonable to aggregate the enhanced image features to point features.} We argue that the proposed CB-Fusion is more effective than the direct fusion in EPNet, as it allows more interaction between the two modalities.
More concretely, the geometric information from LiDAR points is helpful to obtain a more discriminative image representation that is robust to illumination change, and the enhanced image representation in turn brings more plentiful semantics to LiDAR points. \zliu{Besides, CB-Fusion is not limited to point-based detectors~\cite{shi2019pointrcnn,qi2019deep}, but can also be applied to popular voxel-based~\cite{yan2018second,deng2021voxel,yin2021center} and transformer-based detectors~\cite{pan20213d} to boost the detection performance.}

Besides studying multi-sensor fusion at the feature level, we propose a training strategy for further exploring the complementarity of point clouds and images. Our assumption is that the prediction results based on a point cloud should be consistent with those based on its corresponding camera image. Specifically, given a point that is predicted to be from a foreground object, its projection on an image should be predicted to be from the same object as well. Based on this observation,
we adopt the proposed \emph{Multi-Modal Consistency loss}~(MC loss) to guarantee that the network outputs based on different modalities are consistent with each other during the training process.

Finally, in this paper, we emphasize a critical but easily overlooked problem: handling 3D object detection in highly sparse point cloud scenes, which might be an effective strategy to reduce the cost of the LiDAR sensor. Extension experiments demonstrate that our EPNet++ significantly outperforms these existing SOTA methods in highly sparse point cloud cases with the aid of the novel CB-Fusion module and effective MC loss.

In summary, the main contributions in EPNet++ are as follows: 
1) The proposed CB-Fusion module establishes the point-wise feature correspondence between two modalities at different levels of resolution and achieves the cascade bidirectional feature interaction, leading to more comprehensive feature representations for 3D detection tasks. \zliu{Besides, CB-Fusion shows superior generalization capability to multiple architectures, including point-based~\cite{shi2019pointrcnn}, voxel-based~\cite{yan2018second,deng2021voxel,yin2021center} and transformer-based~\cite{pan20213d} detectors.}
2) \zliu{The MC loss encourages the consistency of the predicted confidences of two modalities, which can alleviate the ambiguity, especially when there is a huge gap between the confidence scores of two modalities.}
3) The proposed EPNet++ achieves the competitive even SOTA results on three common 3D object detection benchmark datasets, \textit{i.e.}, the KITTI dataset~\cite{geiger2012we}, SUN-RGBD dataset~\cite{song2015sun} and JRDB dataset~\cite{martin2021jrdb}.
Moreover, EPNet++ achieves promising performance even for highly sparse point cloud scenes, which alleviates the demanding requirements of expensive high-resolution LiDAR sensors.

\section{Related Works}

\subsection{3D Object Detection with Camera Images} 
Achieving 3D object detection at a lower cost is still an important goal for autonomous driving technology. Compared with LiDAR sensors, the price of camera sensors is much lower. Therefore, current 3D object detection methods pay much attention to camera images, such as monocular~\cite{ma2019accurate,qin2019monogrnet,ku2019monopsr,li2019gs3d,liu2019deep,liu2021autoshape} and stereo images~\cite{licvpr2019,wangcvpr2019,chen2020dsgn}. 
Specifically, Chen \textit{et al.}~\cite{chen2016monocular} obtain 2D bounding boxes with a CNN-based object detector and infer their corresponding 3D bounding boxes with semantic, context, and shape information. Deep3DBox~\cite{mousavian20173d} estimates localization and orientation from 2D bounding boxes of objects by exploiting the constraint of projective geometry. However, these methods based on the camera image have difficulty generating accurate 3D bounding boxes due to the lack of depth information. Compared with these image-based methods, our proposed EPNet++ achieves much better detection performance through properly fusing the camera image and the highly sparse point cloud~(\textit{e.g.}, 2048 points) provided by a LiDAR sensor. 

\subsection{3D Object Detection with Point Clouds}
According to data representations of point clouds, we further divide the 3D object detection with point clouds into voxel-based and point-based methods. 

\vspace{1ex}\noindent\textbf{voxel-based Methods.}
Since point clouds are irregular data formats, some existing methods attempt to convert a point cloud into a regular voxel grid representation and then achieve 3D object detection via 2D/3D convolutional neural network~(CNN). In detail, 
VoxelNet~\cite{zhou2018voxelnet} divides a point cloud into voxels and employs stacked voxel feature encoding layers to extract voxel features. SECOND~\cite{yan2018second} further introduces a sparse convolution operation to improve the computational efficiency of ~\cite{zhou2018voxelnet}. To get rid of time-consuming 3D convolution operations, The following work PointPillars~\cite{lang2019pointpillars} converts the point cloud to a pseudo-image and applies 2D CNN to produce the final detection results. Some other works~\cite{shi2020pv,shi2020part,deng2021voxel,li2021lidar} extract the voxel features via 3D convolutional operations and obtain more accurate predictions in a coarse-to-refine two-stage manner. These methods are effective, but they are sensitive to the parameter of voxel size in terms of detection performance and speed. 

\vspace{1ex}\noindent\textbf{Point-based Methods.}
PointRCNN~\cite{shi2019pointrcnn} is a pioneering point-based two-stage detector, which directly produces 3D proposals based on PointNets~\cite{qi2017pointnet,qi2017pointnet++} from the whole point cloud and refines these coarse proposals via 3D ROI pooling operation on the RCNN stage. Then, STD~\cite{yang2019std} applies a point-based proposal generation to achieve high recall through spherical anchors. The following work 3DSSD~\cite{yang20203dssd} proposes a combination of F-FPS and D-FPS to improve the STD~\cite{yang2019std} in terms of accuracy and efficiency. In addition, VoteNet~\cite{qi2019deep} presents a novel deep hough voting for 3D object detection. These methods can directly process the raw point cloud.
Therefore, to establish the accurate correspondence between the camera image and LiDAR point cloud, we utilize the popular point-based detector  PointRCNN~\cite{shi2019pointrcnn} as our backbone in this paper.



\subsection{3D Object Detection with Multiple Sensors} 
Recently, much progress has been made in exploiting multiple sensors, such as camera images and LiDAR point clouds. 
We roughly divide these existing fusion methods into three categories: multi-view fusion, voxel-based and image fusion, and point-based and image fusion.  

\vspace{1ex}\noindent\textbf{Multi-View Fusion Methods.}
As the information of a single view~(\textit{e.g.}, the front view image or BEV point cloud) is usually not sufficient for understanding real scenes, some researchers try to explore multi-view fusion to improve the performance of 3D object detection tasks. MV3D~\cite{chen2017multi} and AVOD~\cite{ku2018joint} refine the detection box via fusing BEV and camera feature maps for each ROI region. 
Although these multi-view approaches usually outperform single-view-based methods, they still suffer from information loss due to the process of converting point clouds to a specific view. In contrast, our proposed fusion module directly operates on the LiDAR point cloud and thus better retains the geometric structure information.

\vspace{1ex}\noindent\textbf{Voxel $\&$ Image Fusion Methods.}
Many recent LiDAR-only methods convert the raw LiDAR point cloud to regular voxel grids for 3D object detection, thanks to its effectiveness and efficiency. To further improve the robustness of 3D detectors, some researchers~\cite{liang2018deep,sindagi2019mvx,zhang2020multi,yoo20203d} devote their efforts to the voxel-based and camera image fusion methods. 
Specifically, ConFuse~\cite{liang2018deep} proposes a novel continuous fusion layer, which not only achieves the voxel-wise alignment between BEV and image feature maps but also captures local information to improve the detection performance. MVX-Net~\cite{sindagi2019mvx} proposes to enhance the voxel feature representations with semantic image features by fusing the features of the camera image and LiDAR point cloud in the early stage. 3D-CVF~\cite{yoo20203d} effectively fuses the spatial feature from both camera image and LiDAR point cloud via utilizing a cross-view spatial feature fusion strategy. 
Due to the quantized error brought from the voxelization operation, 
these methods have limitations in establishing the accurate corresponding relationship between the camera image and the LiDAR point cloud. 
Compared with them, our approach achieves fine-grained correspondence by directly dealing with raw point clouds.

\vspace{1ex}\noindent\textbf{Raw Point Cloud $\&$ Image Fusion Methods.}
Considering that the point cloud possesses rich geometric structure information but lacks plentiful semantic information, some researchers~\cite{xu2018pointfusion,vora2020pointpainting,huang2020epnet} have tried to fuse the raw point cloud and the camera image. Specifically,
PointFusion~\cite{xu2018pointfusion} and SIFRNet~\cite{zhao20193d} first extract semantic features and produce 2D proposals from camera images using off-the-shelf 2D detectors~\cite{liu2016ssd,redmon2016you,ren2015faster,he2017mask}. The extracted semantic features are then combined with the point features extracted from the corresponding frustum to generate 3D bounding boxes.
PointPainting~\cite{vora2020pointpainting} enriches each point feature with the corresponding output class scores predicted by a pre-trained image semantic segmentation network. PI-RCNN~\cite{xie2020pi} employs a segmentation sub-network to extract full-resolution semantic feature maps from images and then fuses the multi-sensor features via a powerful PACF attention module. ImVoteNet~\cite{qi2020imvotenet} further improves the detection performance of VoteNet~\cite{qi2019deep} by lifting 2D camera images votes as well as geometric, semantic, and texture cues from an off-the-shelf 2D detector and then combining them with 3D votes in point clouds. \zliu{DenseFusion~\cite{wang2019densefusion} combines cropped images of interesting objects and corresponding masked point clouds at an object level in a dense pixel-wise fusion, with the help of a  segmentation network~\cite{xiang2017posecnn}.}
Although these methods are effective, their performance still depends on pre-trained 2D detectors or 2D image segmentation networks. Compared with them,
our preliminary work EPNet~\cite{huang2020epnet} enhances the semantic information of point features with image features at different levels of resolution in a point-wise manner, which is fully end-to-end trainable.
In this paper, we further extend EPNet~\cite{huang2020epnet} to EPNet++, which leverages a cascade bi-directional fusion paradigm to allow more information exchange between two modalities and obtain a more comprehensive and robust point feature representation for the subsequent estimation of 3D boxes.


\begin{figure*}[htbp]
\centering
  \includegraphics[width=0.96\linewidth]{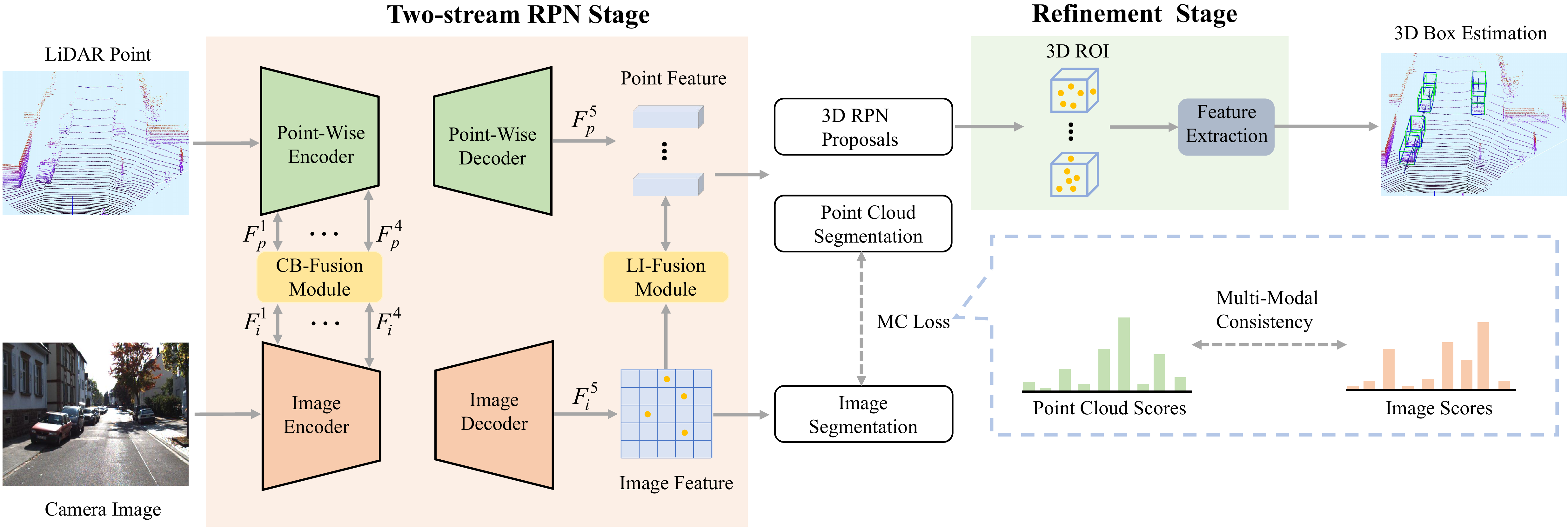}
\caption{Illustration of the overall pipeline of EPNet++ for 3D object detection, which includes a two-steam RPN stage and a refinement stage. In the two-stream RPN stage, several \emph{Cascade Bi-directional Fusion}~(CB-Fusion) modules are employed to establish bi-directional information exchange pathways between two streams in multiple scales, leading to enhanced feature representations. Besides, we utilize the \emph{Multi-Modal Consistency loss}~(MC loss) to encourage the consistency of the segmentation confidences predicted by these two streams, which is helpful for selecting high-quality 3D boxes. Finally, the high-quality 3D proposals are fed into the refinement network to produce the final detection results.}
\label{method_our_architecture}
\end{figure*}

\section{Methods}

Exploiting the complementary information of multiple sensors is crucial for improving the accuracy and robustness of 3D object detectors, especially for sparse scenes. In this paper, we design a novel multi-modal fusion framework for 3D object detection named EPNet++, which is illustrated in Fig.~\ref{method_our_architecture}. 
EPNet++ is trainable in an end-to-end manner, which involves a two-stream RPN for 3D proposal generation and a refinement network to produce more precise 3D bounding boxes.
To effectively explore the complementarity of different sensors, we design the fusion mechanism from both the feature representations and loss constraints. 
First, we propose two feature fusion modules, including a \emph{LiDAR-guided Image Fusion}~(LI-Fusion) module to unidirectionally enhance the semantic information of point features with image features and a \emph{Cascade Bi-directional Fusion}~(CB-Fusion) module to enable bidirectional feature enhancement between LiDAR and camera images.
Second, we employ the \emph{Multi-Modal Consistency loss}~(MC loss) to promote consistency of the foreground/background classification confidence predicted from the LiDAR point cloud and camera image. 
In the following, we present the technical details of EPNet++.



\subsection{Two-Steam RPN} \label{method_rpn}

As shown in Fig.~\ref{method_our_architecture}, the two-stream RPN consists of an image stream and a geometric stream, each of which includes an encoder and a decoder. These two streams can be seamlessly bridged through the proposed LI-Fusion/CB-Fusion modules. Concretely, we employ several CB-Fusion modules in the encoder at multiple scales for early fusion, which enables instant and effective information exchanges between the two streams and produces enhanced features with richer semantic and geometry information. \zliu{Note that we do not employ CB-Fusion modules in the decoder for two reasons. First, multiple transposed convolution layers~(resp.~Feature Propogation layers~\cite{qi2017pointnet++}) are used in the image stream~(resp. geometric stream)  to recover the resolution, which further leads to more serious misalignment. Second, the architecture with the CB-Fusion removed from the decoder is more efficient and thus friendly for deployment.}
Besides, at the end of the two-stream RPN, we simply adopt an LI-Fusion module to enhance the point features for the following detection task. Although CB-Fusion is also an alternative, we prefer LI-Fusion here for its efficiency when processing image feature maps with high resolution. 
In the following,  we will present more details about each component of the two-stream RPN.

\subsubsection{Image Stream}
\label{method:img_stream}
The image stream takes camera images as input and extracts the semantic image information with a set of convolution operations. We adopt a simple structure  
composed of an image encoder and an image decoder, which is illustrated at the bottom of Fig.~\ref{method_our_architecture}. The image encoder contains four light-weighted convolutional blocks. Each convolutional block consists of two $3\times3$ convolution layers followed by a batch normalization layer~\cite{ioffe2015batch} and a ReLU activation function. We set the second convolution layer in each block with a stride of 2 to enlarge the receptive field and save GPU memory. $F_i^k$~($k$=1, 2, 3, 4) denotes the outputs of these four convolutional blocks. $F_i^k$ provides sufficient semantic image information to enrich the LiDAR point features in different scales. The image decoder includes four parallel transposed convolution layers with different strides to recover the image resolution, leading to feature maps of the same resolution as the original image. Then, we combine them in a concatenation manner and obtain a more representative feature map $F_i^5$, which contains rich image semantic information. As is shown later, the feature map $F_i^5$ is also employed to enhance the LiDAR point features to generate more reliable 3D proposals. 

\subsubsection{Geometric Stream}

As shown in the top branch of Fig.~\ref{method_our_architecture}, the geometric stream takes LiDAR point cloud as input and generates 3D proposals for the subsequent refinement stage. We take the popular  PointNet++~\cite{qi2017pointnet++} as the backbone of the geometric stream, which is composed of a point-wise encoder for feature aggregation and a point-wise decoder for feature restoration. The encoder consists of four Set Abstraction~(SA) layers and the decoder is composed of four corresponding Feature Propogation~(FP) layers.
For the convenience of description, the outputs of four SA layers and the last FP layer are denoted as $F_p^k$~($k$=1, 2, 3, 4) and $F_p^5$, respectively. We combine the point features $F_p^k$ with the corresponding semantic image features $F_i^k$ with the aid of our CB-Fusion module. Besides,
we further enrich the decoded point feature $F_p^5$ with the high-resolution image feature $F_i^5$ through an LI-Fusion module to obtain a compact and discriminative feature representation, which is then fed to the RPN head for foreground point segmentation and 3D proposal generation.

\subsubsection{LI-Fusion Module}\label{method_subsubsec_li_fusion}


\begin{figure*}[htbp]
\centering
  \includegraphics[width=0.96\linewidth]{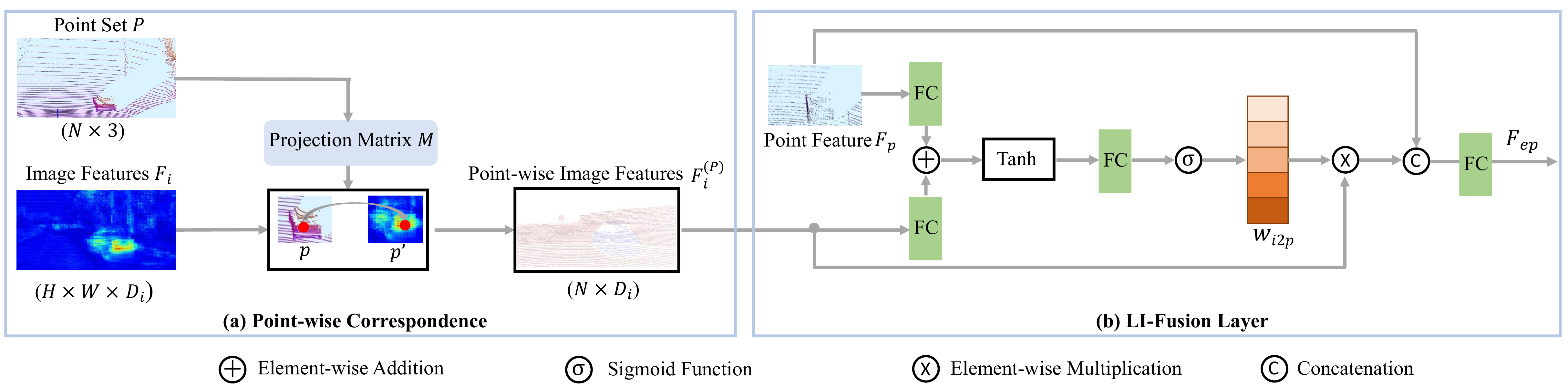}
\caption{Illustration of the structure of the LI-Fusion module. FC represents the fully connected layer.}
\label{fig_li_fusion}
\end{figure*}


The LI-Fusion module aims at enhancing the point cloud feature with rich semantic information from the image stream. As is illustrated in Fig.~\ref{fig_li_fusion}, the LI-Fusion module involves two core components, \textit{i.e.}, a point-wise correspondence operation and an LI-Fusion layer. 


\noindent\zliu{\textbf{Point-wise Correspondence} establishes the point-wise correspondence between 2D location and 3D point cloud through a known projection matrix $M$. For a particular point $p$ in the point set $P$, its corresponding position in the camera image is $p'= Mp$.  Then we use bilinear interpolation to get the corresponding image feature at the continuous coordinates, which well leverages the neighboring context information around the projected position.} This process can be formulated as follows:
\begin{equation}
F_{i}^{(p)}=\mathcal{B}(F_i^{(\mathcal{N}(p'))}),
\end{equation}
where $F_{i}^{(p)}$ is the corresponding image feature for a specific point $p$, $\mathcal{B}$ denotes the bilinear interpolation function, and $F_i^{(\mathcal{N}(p'))}$ represents the image features of the neighboring pixels for the sampling position $p'$. Finally, we can obtain  the total point-wise image features $F_i^{(P)}$ for a point set $P$.

\noindent\zliu{\textbf{LI-Fusion Layer.}~Although concatenation is a simple and common method for feature fusion, it introduces noisy and harmful information when the raw data deteriorates. For example,  the sensor outputs low-quality camera images when it is under bad conditions, such as bad illumination, occlusion, extreme weather~(rainy and snowy), and many other extreme conditions. To overcome this issue, we design an attention mechanism to 
adaptively attend to reliable features while suppressing harmful ones by computing the relevance of image and point features.} As illustrated in Fig.~\ref{fig_li_fusion}~(b), we first feed the point cloud feature $F_p$ and the point-wise image feature $F_i^{(P)}$ into a fully connected layer and map them into the same channel dimension. Then we add them together to form a compact feature representation, which is then compressed into a weight map with a single channel through another fully connected layer. Subsequently, a sigmoid activation function is used to normalize the weight map into the range of [0, 1]. Thus, we can obtain the \emph{Image-to-Point}~(I2P) attention gate weight $w_{i2p}$, which indicates the relevance between the image feature and the point feature.  $w_{i2p}$ can be calculated as follows:
\begin{equation}
\label{formula_w_i2p}
    w_{i2p} = \sigma (W_1\tanh(W_2 F_p + W_3 F_i^{(P)})),
\end{equation}
where $W_1$, $W_2$, $W_3$ denote the learnable weight matrices in our LI-Fusion layer, and $\sigma$ represents the sigmoid function.


After obtaining the I2P attention gate weight $w_{i2p}$, we combine the point cloud feature $F_p$ and the point-wise semantic image feature $F_i^{(P)}$ in a concatenation manner, which can be formularized as follows:
\begin{equation}
    F_{p}' = F_p \ || \ w_{i2p} \cdot F_i^{(P)}.
\end{equation}
Finally, to make the LI-Fusion module more flexible, we compress the channel dimension of the combined feature $F_{p}'$ to be the same as the input point cloud feature $F_p$ via a fully connected layer and obtain the enhanced point cloud feature $F_{ep}$.

\subsubsection{CB-Fusion Module}
\begin{figure}[htbp]
\centering
  \includegraphics[width=0.96\linewidth]{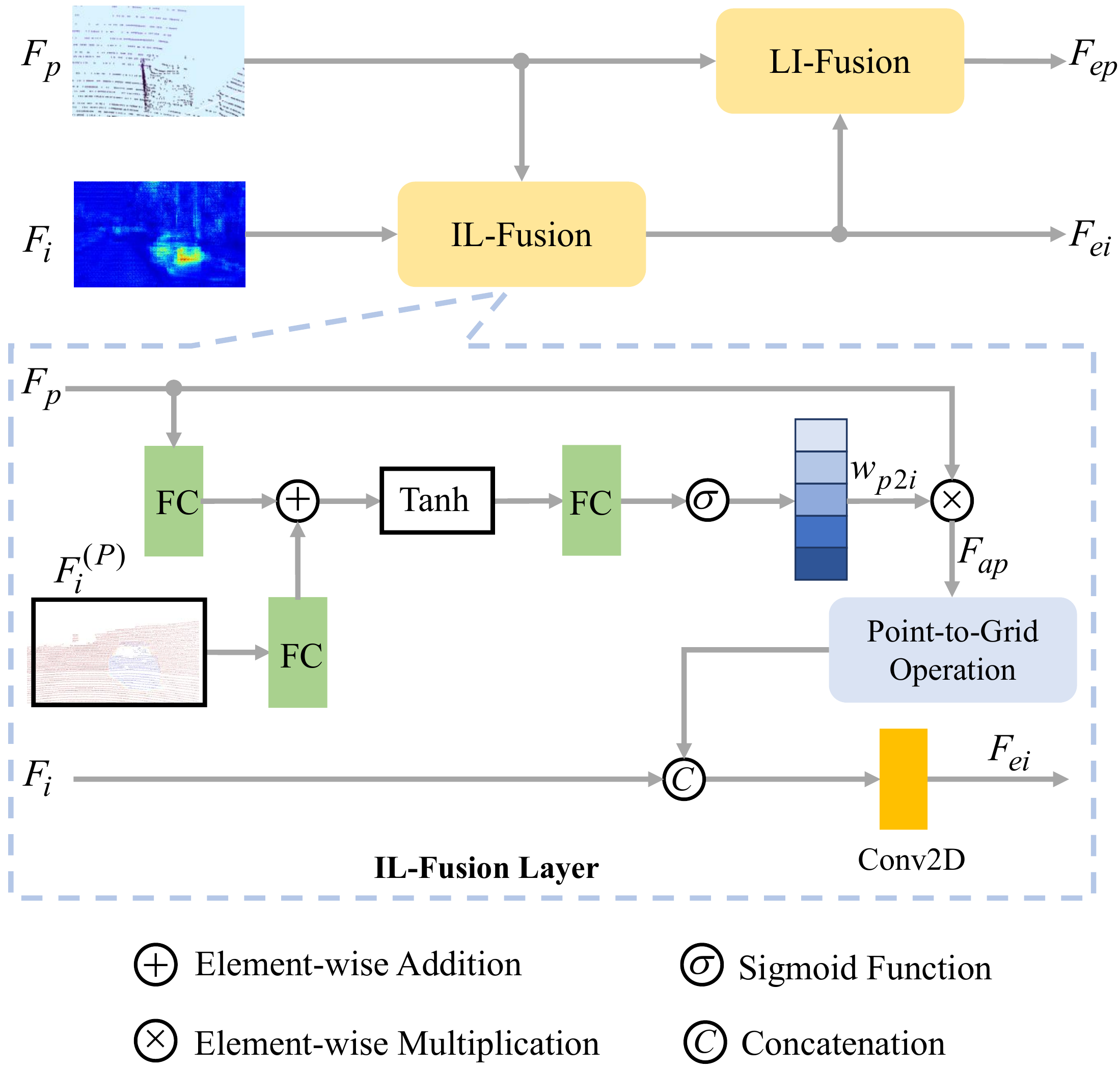}
\caption{Illustration of the structure of the CB-Fusion module, which enables a bi-directional fusion pathway and enhances the image features and the point features in sequential order. }
\label{fig_cb_fusion}
\end{figure}

The above LI-Fusion module only utilizes the image feature to enhance the semantics information of the point cloud.
Although effective, the LI-Fusion module only achieves a unidirectional fusion pathway from image to point cloud, which may not sufficiently exploit the complementarity between these two modalities. We argue that the reversed pathway from the point cloud to the image is also valuable. The reason is twofold. First, the geometric information from the LiDAR sensor is beneficial for understanding the object contour information in the camera images. Second, the depth information collected from a LiDAR sensor is more robust to the change of illumination and provides cues for better localizing objects in the camera images. Motivated by these observations, we propose an \emph{Image-guided LiDAR Fusion}~(IL-Fusion) module to enable the feature enhancement pathway from LiDAR to image. The combination of IL-Fusion and LI-Fusion forms our novel \emph{Cascaded Bi-directional Multi-Modal Fusion}~(CB-Fusion) module. As is shown in Fig.~\ref{fig_cb_fusion}, \zliu{the CB-Fusion module first generates more informative image features with point cloud depth information embedded through IL-Fusion, and then outputs a more robust point cloud feature fused with image semantic information through LI-Fusion.} In addition to employing the same point-wise correspondence operation mentioned in the part of the LI-Fusion module to obtain the point-wise image feature $F_i^{(P)}$, the IL-Fusion module involves an IL-Fusion layer, which is detailed in the following.



\noindent\textbf{IL-Fusion layer.}~\zliu{Similar to LI-Fusion, IL-Fusion is also designed to improve the robustness of fusion, when the LiDAR sensor is under bad conditions, such as extreme weather~(rainy and snowy), etc.} To alleviate this issue, we introduce
the \emph{Point-to-Image}~(P2I) attention gate weight $w_{p2i}$ to adaptively compute the importance of the point cloud features to the image features. As shown in the bottom of Fig.~\ref{fig_cb_fusion}, the P2I attention gate weight $w_{p2i}$ is calculated in a similar mechanism with the I2P attention gate weight $w_{i2p}$ by only swapping the input features of $F_p$ and $F_i^{(P)}$ in the formula~(\ref{formula_w_i2p}). 
Subsequently, we multiply the point cloud feature $F_p$ with $w_{p2i}$ and obtain the point cloud feature with attention enhancement $F_{ap}$.



To combine the enhanced point feature $F_{ap}$ of shape $N \times D_p$ and the image feature $F_i$ of shape $H\times W \times D_i$, we need to convert them into the same shape. Here, $D_p$ and $D_i$ represent the channel dimension for the features of $F_{ap}$ and $F_i$, respectively.
The point-to-grid operation scatters the point features into the image grid based on the point-wise correspondence and produces the projected point feature with the same shape of $F_i$. However, the projected location of a point is rarely on the image grid and usually between pixels. 
Thus, we restore the feature with an integer grid by applying bilinear interpolation to its adjacent point features, leading to a grid-like feature map.        

After converting $F_{ap}$ into the grid-like point feature map, we combine it with the image feature $F_i$ in a concatenation manner. Finally, we feed the combined features to a 2D convolution layer with the kernel size of 3 and the stride of 1, leading to the enhanced image feature $F_{ei}$.

\subsection{Refinement Stage} \label{method_refinment}
As shown in Fig.~\ref{method_our_architecture}, we employ the NMS procedure to keep the high-quality proposals predicted by the two-stream RPN and feed them into the refinement network. Similar to PointRCNN~\cite{shi2019pointrcnn}, we randomly select 512 points inside a proposal as its 3D RoI feature descriptor.
For those proposals with less than 512 points, we simply sample the points with replacements. The refinement network includes three SA layers~\cite{qi2017pointnet++} and two detection heads consisting of two cascaded $1\times 1$ convolution layers. Specifically, three SA layers are used to extract a compact global descriptor for each 3D ROI, and two detection heads are employed to classify and regress the final 3D objects. 


\subsection{Multi-Modal Consistency Loss} \label{method_mc_loss}

\begin{figure}[htbp]
\centering
  \includegraphics[width=0.96\linewidth]{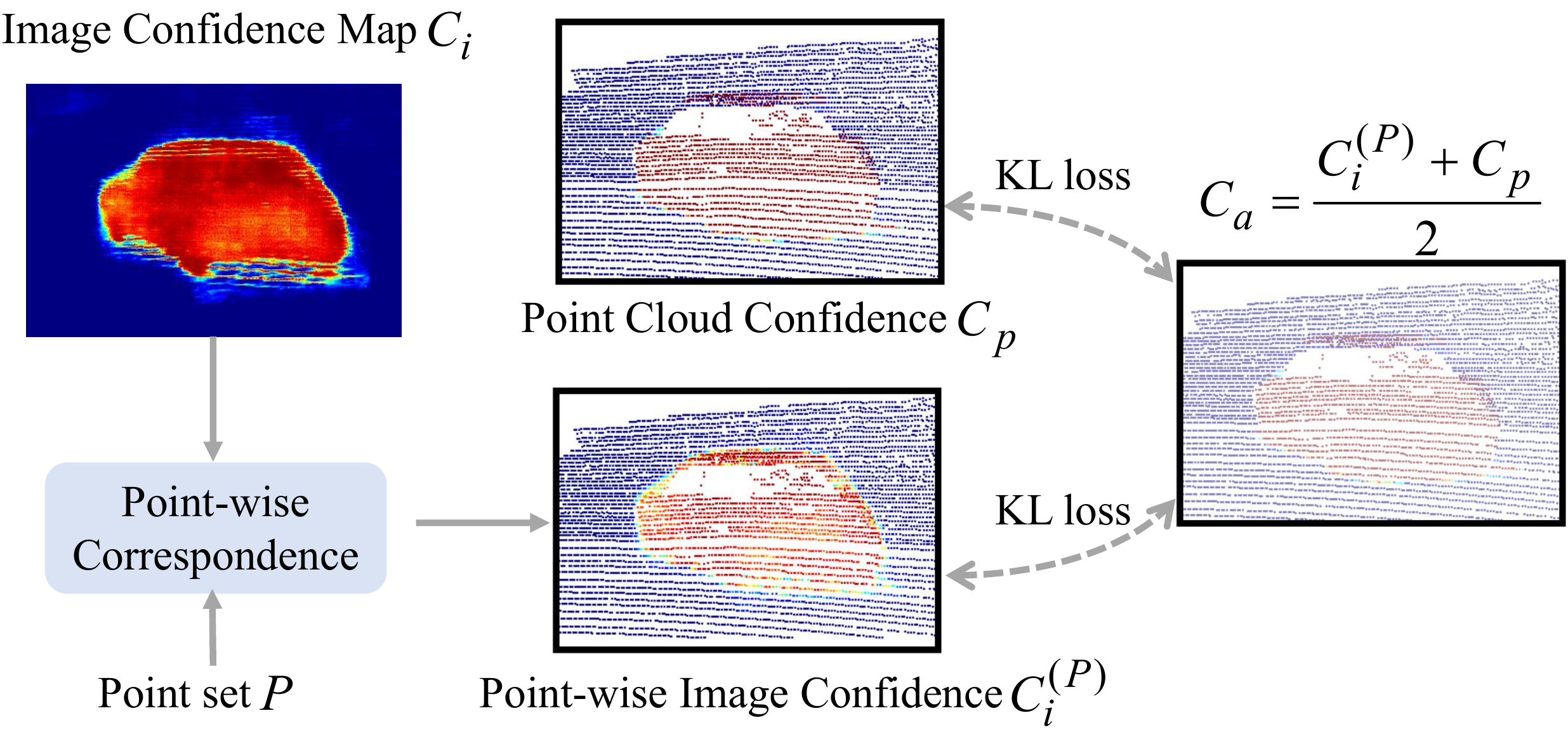}
\caption{Illustration of the process of the MC loss.}
\label{mc_loss}
\end{figure}

The two-stream RPN in EPNet++ generates a corresponding proposal for each foreground point, which is similar to PointRCNN~\cite{shi2019pointrcnn}. In this process, how to obtain reliable confidence scores of foreground points is a crucial role in selecting high-quality boxes in the process of RPN. However, the image stream and the geometric stream might produce distinct confidence for the foreground object prediction. 
Therefore, it indicates that there is some room for improvement by properly using the valuable confidence difference.
For example, compared with the predicted image scores from the image stream, the scores from the geometric stream may be more uncertain for recognizing the foreground or background in a highly sparse point cloud scene. However, in a denser case, the point cloud scores from the geometric stream are more reliable in dealing with occluded objects and distinguishing the boundary of objects than the camera image scores. Thus, it naturally motivates us to leverage the complementarity of different modalities by aligning the prediction confidence of the two streams.

We propose the \textit{Multi-Modal Consistency}~(MC) loss to guarantee the consistency between the confidence scores predicted by the two streams. Concretely, we adopt a simple yet effective strategy, that is, pulling the confidence scores of both streams to their average value. Fig.~\ref{mc_loss} illustrates the details of our MC loss. Given a certain point from the input point set $P$, we find its corresponding pixel from the image confidence map $C_i$ through the point-wise correspondence process. $C_p$ and $C_i^{(P)}$ denote the confidence of the LiDAR points and the corresponding pixels. 
MC loss pulls both $C_p$ and $C_i^{(P)}$ to their average value $C_a = (C_i^{(P)} + C_p)/{2}$ through KL loss~\cite{he2019bounding}. \zliu{The averaged confidence score can avoid the overconfident prediction from one of the modalities. Since when there is a huge gap between the confidence scores of two modalities, this may bring ambiguity in judging whether an object actually exists.} Besides, when both $C_p$ and $C_i^{(P)}$ are smaller than a predefined threshold score $\tau$ for a specific spatial position, we regard it as a background point, which is not calculated in KL loss. Finally, the MC loss can be formularized as follows:

\begin{equation}
\mathbb{I}_{\mathrm{m}}=\left\{
\begin{array}{cl}
1 &  \mathrm{Max}(C_i^{(P)}, C_p) > \tau  \\
0 &  \mathrm{Max}{(C_i^{(P)}, C_p) \le \tau} \\
\end{array} \right.
\end{equation}
\begin{equation}
   L_{i2p} = \mathbb{I}_{\mathrm{m}} \cdot KL(C_i^{(P)} || C_a) ,
 \end{equation}
\begin{equation} 
    L_{p2i} = \mathbb{I}_{\mathrm{m}} \cdot KL(C_p || C_a) ,
 \end{equation}
\begin{equation}
   L_{mc} = \frac{1}{N}\sum_{j=0}^{N} (\lambda_1 L_{i2p}^{(j)} + \lambda_2 L_{p2i}^{(j)}),
\end{equation}
where $\mathbb{I}_{\mathrm{m}}$ is an indicator function, which outputs ones for foreground positions while zeros for background positions. $\tau$ is the score threshold for differentiating foreground and background. $L_{i2p}$ represents the MC loss for pulling the $C_i^{(P)}$ to $C_a$, while $L_{p2i}$ pulls $C_p$.
$N$ is the total number of input points. $\lambda_1$ and $\lambda_2$ represent the balancing weights of $L_{i2p}$ and  $L_{p2i}$. More discussions about the effects of $\lambda_1$ and $\lambda_2$ on the detection performance are in Section~\ref{exp_lambda}.

\subsection{Total Loss Function}\label{method_overall_loss}

We utilize a multi-task loss function for jointly optimizing the two-stream RPN and the refinement network. The total loss can be formulated as:
\begin{equation}
L_{total} = L_{rpn} +  L_{rcnn},
\end{equation}
where $L_{rpn}$ and $L_{rcnn}$ denote the training objectives for the two-stream RPN and the refinement network, both of which adopt a similar optimizing goal, including a classification loss, a regression loss and a \emph{Consistency Enforcing loss}~(CE loss)~\cite{huang2020epnet}. More concretely, we adopt the focal loss~\cite{lin2017focal} as our classification loss to balance the positive and negative samples with the setting of $\alpha=0.25$ and $\gamma=2.0$. For a bounding box, the network needs to regress its center point~$(x, y, z)$, size~$(l, h, w)$, and orientation $\theta$.
The CE loss is employed to encourage the consistency of the classification and regression confidence scores, which was first proposed in our preliminary work EPNet~\cite{huang2020epnet}. As the effectiveness of CE loss for improving the detection performance has been verified, we adopt the CE loss $L_{ce}$ as the default option in our EPNet++. The details of CE loss will be introduced at the end of this part.

Since the range of the Y-axis~(the vertical axis) is relatively small, we directly calculate its offset to the ground truth with a smooth L1 loss~\cite{girshick2015fast}. Similarly, the size of the bounding box~$(h, w,l)$ is also optimized with a smooth L1 loss. As for the X-axis, the Z-axis and the orientation $\theta$, we adopt a bin-based regression loss~\cite{shi2019pointrcnn,qi2018frustum}. For each foreground point, we split its neighboring area into several bins. The bin-based loss first predicts which bin
$b_u$ the center point falls in, and then regresses the residual offset $r_u$ within the bin. In addition to the common loss mentioned above, the two-stream RPN loss includes an image segmentation loss $L_{ims}$ and an MC loss $L_{mc}$. Finally, we formulate the loss functions in the two-stream RPN stage as follows:
\begin{equation} \label{rpn_loss}
L_{rpn}= L_{cls} + L_{reg}  + L_{ims} + L_{mc} + \beta L_{ce},
\end{equation}
\begin{equation}
    L_{cls} = -\alpha(1-C_p)^\gamma \log C_p, 
\end{equation}
\begin{equation}
    L_{ims} = -\alpha(1-C_i)^\gamma \log C_i,
\end{equation}
\begin{equation}
    L_{reg} = \sum_{u \in {x,z, \theta}}E(b_u, \hat{b_u}) + \sum_{u \in {x,y,z,h,w,l,\theta}} S(r_u, \hat{r_u}),
\end{equation}
where $E$ and $S$ denote the cross-entropy loss and the smooth L1 loss, respectively. $\beta$ is the  balance  weight for CE loss.
$\hat{b_u}$ and $\hat{r_u}$ denote the ground truth of the bins and the residual offsets. Similarly, $L_{rcnn}$ can be computed by only removing $L_{mc}$ and $L_{ims}$  in the formula (\ref{rpn_loss}).

\noindent\textbf{CE loss.}~In general, we assume that the classification confidence can serve as an agent for the real IoU between the bounding and the ground truth, \textit{i.e.}, the localization confidence. However, the classification confidence and the localization confidence are often inconsistent, leading to sub-optimal performance. Thus, we use a CE loss to ensure the consistency between the localization and classification confidence so that boxes with high localization confidence possess high classification confidence and vice versa. The CE loss can be written as follows:
\begin{equation}
L_{ce} = -log(C_c \cdot \frac{Area(B_p \cap B_g)}{Area(B_p \cup B_g)})
\end{equation}
where $B_p$ and $B_g$ represent the predicted bounding box and the ground truth. Area($\star$) means to calculate the area of $\star$. And $C_c$ denotes the classification confidence. In particular, $C_c$ is equal to $C_a$ when MC loss is enabled, otherwise $C_c=C_p$. Towards optimizing this loss function, the classification confidence and localization confidence~(\textit{i.e.}, the IoU) are encouraged to be as high as possible jointly. Hence, boxes with large overlaps will possess high classification possibilities and be kept in the NMS procedure.

\section{Experiments}

\subsection{Experimental Datasets and Evaluation Metrics} \label{exp_setup}

\begin{table*}[htbp]
\scriptsize
\centering
\caption{Quantitative comparisons with the state-of-the-art 3D object detection methods on the KITTI test benchmark. For \emph{Modality} column, \emph{P} and \emph{I} denote the point cloud and the camera image, respectively. The same denotations are used in the following  Table~\ref{sunrgbd_results}, Table~\ref{jrdb_reults} and Table~\ref{sparse_cmp_results}. }
\begin{tabular}{l|c|c|c|c|c|c|c|c|c|c|c}
\hline
\multirow{2}{*}{Method} 
&\multirow{2}{*}{Conference} 
&\multirow{2}{*}{Modality} 
&\multicolumn{3}{c|}{Cars(Recall40)} 
&\multicolumn{3}{c|}{Pedestrians(Recall40)} 
&\multicolumn{3}{c}{Cyclists(Recall40)}\\
\cline{4-12}
& & &Easy & Moderate & Hard & Easy & Moderate & Hard & Easy & Moderate & Hard\\
\hline
\hline
MonoGRNet~\cite{qin2021monogrnet} &TPAMI 2021 &\multirow{9}{*}{I} &9.61 &5.74 &4.25 &- &- &- &- &- &-  \\
M3D-RPN~\cite{brazil2019m3d} &ICCV 2019 & &14.76 &9.71 &7.42 &4.92 &3.48 &2.94  &0.94 &0.65 &0.47 \\
MonoPair~\cite{chen2020monopair} &CVPR 2020 & &13.04 &9.99 &8.65 &10.02 &6.68 &5.53   &3.79 &2.12 &1.83  \\
RTM3D~\cite{li2020rtm3d} &ECCV 2020 & &14.41 &10.34 &8.77 &- &- &- &- &- &-  \\
PatchNet~\cite{ma2020rethinking} &ECCV 2020 & &15.68 &11.12 &10.17 &- &- &- &- &- &-  \\
CaDDN~\cite{reading2021categorical} &CVPR 2021 & &19.17 &13.41 &11.46 &12.87 &8.14 &6.76 &7.00 &3.41 &3.30  \\
GUP Net~\cite{lu2021geometry} &ICCV 2021 & &20.11 &14.20 &11.77 &14.72 &9.53 &7.87 &4.18 &2.65 &2.09  \\
DD3d~\cite{park2021pseudo} &ICCV 2021 & &23.22 &16.34 &14.20 &- &- &- &- &- &-  \\
\hline
\hline
PointPillars~\cite{lang2019pointpillars} &CVPR 2019 &\multirow{22}{*}{P}  &82.58 &74.31 &68.99 &51.45 &41.92 &38.89 &77.10 &58.65 &51.92 \\
PointRCNN~\cite{shi2019pointrcnn} &CVPR 2019 & &86.96 &75.64 &70.70 &47.98 &39.37 &36.01 &74.96 &58.82 &52.53 \\
TANet~\cite{liu2020tanet} &AAAI 2020 & &84.39 &75.94 &68.82 &53.72 &44.34 &40.49 &75.70 &59.44 &52.53   \\
Pointformer~\cite{pan20213d} &CVPR 2021 & &87.13 &77.06 &69.25 &50.67 &42.43 &39.60 &75.01 &59.80 &53.99  \\
Fast PointRCNN~\cite{chen2019fast} &ICCV 2019 &  &85.29 &77.40 &70.24 &- &- &- &- &- &-  \\
HotSpotNet~\cite{chen2020object} &ECCV 2020 &  &87.60 &78.31 &73.34 &53.10 &\textbf{45.37} &\textbf{41.47} &\textbf{82.59} &\textbf{65.95} &\textbf{59.00} \\
Part$A^{2}$~\cite{shi2020part} &TPAMI 2020 & &87.81 &78.49 &73.51 &53.10 &43.35 &40.06 &79.17 &63.52 &56.93 \\
SERCNN~\cite{zhou2020joint} &CVPR 2020 & &87.74 &78.96 &74.30 &- &- &- &- &- &- \\
SECOND~\cite{yan2018second} &Sensors 2018 &  &87.44 &79.46 &73.97 &- &- &- &- &- &-  \\
Point-GNN~\cite{shi2020pointgnn} &CVPR 2020 &  &88.33 &79.47 &72.29 &51.92 &43.77 &40.14 &78.60 &63.48 &57.08 \\
MGAF-3DSSD~\cite{Anchor-free-MGA-SSD} &ACM MM 2021 &  &88.16 &79.68 &72.39 &- &- &- &- &- &- \\
3DSSD~\cite{yang2019std} &CVPR 2020 &  &88.36 &79.57 &74.55 &\textbf{54.64} &44.27 &40.23 &82.48 &64.10 &56.90 \\
STD~\cite{yang2019std} &ICCV 2019&  &87.95 &79.71 &75.09 &53.29 &42.47 &38.35 &78.69 &61.59 &55.30 \\
SA-SSD~\cite{he2020sassd}  &CVPR 2020 &  &88.75 &79.79 &74.16 &- &- &- &- &- &- \\
CIA-SSD~\cite{zheng2020ciassd}  &AAAI 2021 & &89.59 &80.28 &72.87 &- &- &- &- &- &-  \\
PV-RCNN~\cite{shi2020pv} &CVPR 2020 & &90.25 &81.43 &76.82 &52.17 &43.29 &40.29 &78.60 &63.71 &57.65 \\
FromVoxelToPoint~\cite{FromVoxelToPoint} &ACM 2021&	&88.53 &81.58 &\textbf{77.37} &- &- &- &- &- &-  \\
BANet~\cite{qian2021boundaryaware} &Arxiv 2021 & &89.28	&81.61 &76.58 &- &- &- &- &- &- \\
Voxel R-CNN~\cite{deng2021voxel} &AAAI 2021 & &\textbf{90.90} &81.62 &77.06 &- &- &- &- &- &-  \\
SIENet~\cite{li2021sienet} &Arxiv 2021 & &88.22 &{81.71} &77.22  &- &- &- &- &- &-   \\
M3DETR~\cite{guan2022m3detr} & WACV 2022 & &90.28 &\textbf{81.73} &76.96 & 45.70 &39.94 &37.66 &\textbf{83.83} &\textbf{66.74} &\textbf{59.03} \\
\hline
\hline
MV3D~\cite{chen2017multi} &CVPR 2017 &\multirow{13}{*}{P + I}   &74.97 &63.63 &54.00 &- &- &- &- &- &- \\
Confuse~\cite{liang2018deep} &ECCV 2018 &   &83.68 &68.78 &61.67 &- &- &- &- &- &- \\
F-Pointnet~\cite{qi2018frustum}	&CVPR 2018	&   &82.19 &69.79 &60.59 &50.53 &42.15 &38.08 &72.27&56.12&49.01  \\
PointPainting~\cite{vora2020pointpainting}	&CVPR 2020 &  &82.11 &71.70 &67.08 &50.32 &40.97 &37.87 &77.63&63.78&55.89 \\
AVOD-FPN~\cite{ku2018joint}		&IROS 2018	&   &83.07 &71.76 &65.73 &50.46 &42.27 &39.04 &63.76&50.55&44.93 \\
PI-RCNN~\cite{xie2020pi}			&AAAI 2020	&   &84.37 &74.82 &70.03 &- &- &- &- &- &- \\
F-Convnet~\cite{wang2019frustum}	&IROS 2019	&   &87.36 &76.39 &66.69 &52.16 &43.38 &38.80  & \textbf{81.98} & \textbf{65.07} & \textbf{56.54}  \\
MMF\cite{liang2019multi}			&CVPR 2019 &   &88.40 &77.43 &70.22 &- &- &- &- &- &- \\
CLOCs\_SecCas~\cite{pang2020clocs}	&IROS 2020 &	&86.38 &78.45 &72.45 &- &- &- &- &- &- \\
MVAF~\cite{wang2020multi}			&Arxiv 2020 &   &87.87 &78.71 &75.48 &- &- &- &- &- &- \\
3D-CVF~\cite{yoo20203d}				&ECCV 2020 &   &89.20 &80.05 &73.11 &- &- &- &- &- &- \\
CLOCs\_PVCas~\cite{pang2020clocs}	&IROS 2020 &	&88.94 &80.67 & \textbf{77.15} &- &- &- &- &- &- \\
\hline
EPNet~\cite{huang2020epnet}			&ECCV 2020 &\multirow{2}{*}{P + I}  	&89.81 &79.28 &74.59 &- &- &- &- &- &- \\
\textbf{EPNet++~(Ours)}  &-&  & \textbf{91.37} & \textbf{81.96} &76.71 & \textbf{52.79} & \textbf{44.38} & \textbf{41.29} &76.15 &59.71 &53.67 \\
\hline
\hline
\end{tabular}
\label{kitti_results}
\end{table*}

\noindent\textbf{KITTI Dataset}~\cite{geiger2012we} is a popular benchmark dataset for autonomous driving, which is collected by a \zliu{64-beam LiDAR sensor} and two camera sensors. The dataset  includes 7,481 training frames and 7,518 testing frames for 3D object detection. Following the same dataset split protocol as~\cite{qi2018frustum,shi2019pointrcnn}, the 7,481 frames are further split into 3,712 frames for training and 3,769 frames for validation. 
In addition, for the image segmentation task, we obtain the mask annotations of Cars, Pedestrians and Cyclists from the KINS dataset~\cite{qi2019amodal}.
In our experiments, we provide the results on both the validation and the testing set for three difficulty levels, \textit{i.e.},
\emph{Easy}, \emph{Moderate}, and \emph{Hard}. Objects are categorized into different difficulty levels according to their sizes, occlusion, and truncation. 
Recently, the KITTI dataset adopts a better evaluation protocol~\cite{simonelli2019disentangling} which computes the mean Average Precision~(mAP) using 40 recall positions instead of 11 as before. We compare our methods with state-of-the-art methods under this new evaluation protocol.


\noindent\textbf{JRDB Dataset}~\cite{martin2021jrdb} is a large-scale multi-modal dataset collected from a social mobile manipulator JackRabbot. The dataset is designed for facilitating perceptual tasks necessary for a robot to understand a scene and human behavior. The dataset provides stereo cylindrical $360^{\circ}$ RGB video streams, continuous 3D point clouds scanned from two 16-channel Velodyne LiDARs, and $360^{\circ}$ spherical image from a fisheye camera. There are 54 sequences collected in both indoor and outdoor environments. 
The sequences are divided into 27 training sequences and 27 testing sequences for the 3D detection task. Following the official suggestion, we select 7 out of 27 training sequences as the validation split. 
JRDB provides both the 2D and 3D bounding box annotations for Pedestrians in the full $360^{\circ}$ scenes, different from the KITTI dataset which annotates the front view scenes.
Note that the 2D dense segmentation annotations for our image stream are not available. However, we can obtain sparse segmentation masks. Specifically, we consider the 2D projection pixels as foreground if the corresponding LiDAR points are inside the 3D bounding boxes, otherwise as background. 
The official evaluation metrics are similar to KITTI, while the 3D IoU threshold for Pedestrians is set to 0.3 instead of 0.5 in KITTI.

\noindent\textbf{SUN-RGBD Dataset}~\cite{song2015sun} is an indoor benchmark dataset for 3D object detection. The dataset is composed of 10,335 images with 700 annotated object categories, including 5,285 images for training and 5,050 images for testing. We report results on the testing set for ten main object categories following previous works~\cite{qi2019deep,qi2020imvotenet} since the number of these object categories on the whole dataset is relatively enormous. Following ~\cite{song2015sun}, we adopt mAP as the evaluation metric to measure the 3D object detection performance by setting the value of the 3D IoU threshold as 0.25.

\begin{table*}[htbp]
	\scriptsize
	\centering
	\caption{Quantitative comparisons with SOTA methods for 3D object detection task on the SUN-RGBD validation set. * means the reproduced results.}
		\begin{tabular}{l|c|c||c|c|c|c|c|c|c|c|c|c||c}
			\hline
			Method & Conference & Modality & bathtub  & bed  & bookshelf & chair & desk & dresser & nightstand & sofa  &  table  & toilet  &mAP \\
			\hline
			MLCVNet~\cite{xie2020mlcvnet} &CVPR 2020 &P &79.2 &85.8 &31.9 &75.8 &26.5 &31.3 &61.5 &66.3 &50.4 &89.1 &59.8 \\
			H3DNet~\cite{zhang2020h3dnet} &ECCV 2020 &P &73.8 &85.6 &31.0 &76.7 &29.6 &33.4 &65.5 &66.5 &50.8 &88.2 &60.1 \\
			HGNet~\cite{chen2020hierarchical} &CVPR 2020 &P &78.0 &84.5 &35.7 &75.2 &\textbf{34.3} &37.6 &61.7 &65.7 &51.6 &91.1 &61.6 \\
			Pointformer~\cite{pan20213d}  &CVPR 2021 &P &80.1 &84.3 &32.0 &76.2 &27.0 &37.4 &64.0 &64.9 &51.5 &92.2 &61.1 \\ 
			Group-Free-3D~\cite{liu2021group} &{ICCV 2021} &P &80.0 &87.8 &32.5 &79.4 &32.6 &36.0 &66.7 &70.0 &\textbf{53.8} &91.1 &63.0 \\
			\hline
			DSS~\cite{song2016deep} &CVPR 2016 &P + I  & 44.2 & 78.8 & 11.9 & 61.2 & 20.5 & 6.4 & 15.4 & 53.5 & 50.3 & 78.9 & 42.1\\
			2d-driven~\cite{lahoud20172d} &ICCV 2017&P + I  & 43.5 & 64.5 & 31.4 & 48.3 & 27.9 & 25.9 & 41.9 & 50.4 & 37.0 & 80.4 & 45.1\\
			COG~\cite{ren2016three} &CVPR 2016 &P + I  & 58.3  & 63.7 & 31.8 & 62.2 & 45.2 & 15.5 & 27.4 & 51.0 & 51.3 & 70.1 & 47.6 \\
			PointFusion~\cite{xu2017pointfusion} &CVPR 2018 &P + I  & 37.3 & 68.6 & 37.7 & 55.1 & 17.2 & 24.0 & 32.3 & 53.8 & 31.0 & 83.8 & 44.1 \\
			F-PointNet~\cite{qi2018frustum} &CVPR 2018&P + I  & 43.3  & 81.1 & 33.3 & 64.2 & 24.7 & 32.0 & 58.1 & 61.1 & 51.1 & 90.9 & 54.0 \\
			EPNet~\cite{huang2020epnet} &ECCV 2020 &P + I  & 75.4 & 85.2 & 35.4 & 75.0  & 26.1   & 31.3  &62.0  & {67.2} &{52.1}  & 88.2   & {59.8} \\
			\hline
			\hline
			VoteNet~\cite{qi2019deep} &ICCV 2019 &P &74.4  &83.0 &28.8 &75.3 &22.0 &29.8 &62.2 &64.0 &47.3 &90.1 & 57.7  \\
			VoteNet$^*$~\cite{qi2019deep} &{ICCV 2019} &P  &75.5 & 85.6 & 31.9 & 77.4  & 24.8   & 27.9  &58.6  & 67.4 &51.1  & 90.5   & 59.1 \\
			+LI-Fusion~(EPNet) &ECCV 2020&P + I  & 73.8  & 86.4  & 35.6 & 80.1  & 27.4   &31.8  &63.0  & 67.3 &54.9 & 88.8 & 60.9 \\
			+CB-Fusion~(Ours) &- &P + I  & \textbf{80.3} & 84.7 & 33.1 & 79.7  & 27.2   & 36.2  &63.1  & 69.9 &53.5  & 86.9   & 61.5 \\
			\hline
			ImVoteNet~\cite{qi2020imvotenet} &CVPR 2020&P + I  & 75.9 & 87.6 & 41.3 & 76.7  & 28.7   & 41.4  &\textbf{69.9}  &70.7 &51.1  & 90.5   & 63.4 \\
			ImVoteNet$^*$~\cite{qi2020imvotenet} &CVPR 2020 &P + I  & 77.3 & 88.2 & 41.4 & 79.5  & 30.8   & 39.9  &68.0  & \textbf{72.4} &51.5  & 91.5   & 64.0 \\
            +LI-Fusion~(EPNet) &ECCV 2020 &P + I  & 72.8 & 88.4 & 47.6 & \textbf{80.9}  & 32.3   & 42.5  &64.5  & 70.8  &53.6  & 93.1   & 64.6 \\
			+CB-Fusion~(Ours) &- &P + I  & 76.3 & \textbf{89.1} &  \textbf{47.1} & 80.2  & 32.5   & \textbf{45.2}  &67.4  & 71.9 &51.3  & \textbf{92.4}  & \textbf{65.3} \\
			\hline 
		\end{tabular}
\label{sunrgbd_results}
\end{table*}

\subsection{Implementation Details} \label{exp_imp_detail}

On the large-range outdoor scenes datasets of JRDB and KITTI, we adopt the identical network architecture for EPNet++, as is shown in Fig.~\ref{method_our_architecture}. 
For the indoor SUN-RGBD dataset, we integrate the proposed CB-Fusion and LI-Fusion modules into different backbones, \textit{e.g.,} VoteNet~\cite{qi2019deep} and ImVoteNet~\cite{qi2020imvotenet}, to verify the generalization capability of our method. We carefully follow their settings on SUN-RGBD referring to the original papers. In the following, we mainly focus on the technical details of our EPNet++. 

\noindent\textbf{Network Settings.}~EPNet++ takes both the LiDAR point cloud and the camera image as inputs. For each 3D scene on the KITTI dataset, the range of the LiDAR point cloud is set to [-40, 40], [-1, 3], [0, 70.4] meters along the X~(right), Y~(down), Z~(forward) axis in the camera coordinate, respectively. On the JRDB dataset, we only limit the horizontal range of the point cloud within a radius of 26 meters, without restricting the vertical direction.
The heading orientation $\theta$ is in the range of [-$\pi$, $\pi$] on both datasets. We subsample 16,384 and 32,768 points from the raw LiDAR point cloud as the input of the geometric stream for the KITTI dataset and the JRDB dataset, respectively. The image stream takes images with a resolution of $1280 \times 384$ as input for the KITTI dataset. While on the JRDB dataset, we scale the original image to a smaller size of $1888 \times 240$ for saving GPU memory. Four set abstraction layers are employed to downsample the input LiDAR point cloud to the resolution of 4096~(8192), 1024~(2048), 256~(512), and 64~(128) on the KITTI~(JRDB) dataset, respectively. Four feature propagation layers are used to gradually recover the original resolution of the point cloud for the following foreground segmentation and 3D proposal generation. Similar to the geometric stream, we use four convolution blocks with the same stride of 2 to downsample the input image. Besides, we employ four parallel transposed convolutions with the strides of 2, 4, 8, and 16 to recover the original image resolution from feature maps of different scales. In the NMS process, we select the top 8000 boxes generated by the two-stream RPN according to the comprehensive confidence score $C_a$. After that, we filter redundant boxes with the IoU threshold of 0.8 and obtain 100 positive candidate boxes, which will be refined by the refinement network.

\noindent\textbf{Training Scheme.}~Our two-stream RPN and refinement network are jointly optimized in an end-to-end manner. The regression loss $L_{reg}$ and the CE loss $L_{ce}$ are only applied to positive candidates, \textit{i.e.}, proposals generated by foreground points in the RPN stage, and boxes sharing the IoU larger than 0.55 with the ground truth in the RCNN stage. The balancing weight $\beta$ for the CE loss is set to 5. For the MC loss $L_{mc}$, we simply set the default values of $\lambda_1$ and $\lambda_2$ to 0.5 and 0.5. The score threshold $\tau$ in the indicator function is set to 0.2. We optimize the model by Adaptive Moment Estimation~(Adam)~\cite{kingma2014adam} with the initial learning rate, weight decay, and momentum factor set to 0.002, 0.001, and 0.9, respectively. We train the model with a batch size of 8 for around 50 epochs on the KITTI dataset and 20 epochs on the JRDB dataset. All the experiments are conducted on a machine with four Titan V GPUs using the popular deep learning framework Pytorch~\cite{paszke2019pytorch}.

\noindent\textbf{Data Augmentation.}~Three common data augmentation strategies are adopted to prevent over-fitting, including rotation, flipping, and scale transformations. First, we randomly rotate the point cloud along the vertical axis within the range of $[-\pi/18,\pi/18]$. Then, the point cloud is randomly flipped along the forward axis. Besides, each ground truth box is randomly scaled following the uniform distribution of $[0.95,1.05]$. It is noted that many LiDAR-based methods utilize the GT Sampling data augmentation~\cite{yan2018second} to improve the diversity of objects, which
samples ground truth boxes from the whole dataset and places them into the raw 3D frames to simulate real scenes with crowded objects. Although effective, this data augmentation needs the prior information of road planes, which is usually difficult to acquire for kinds of real scenes.
In addition, the strategy may lead to the misalignment of the camera image and point cloud, which is not so suitable for multi-modal fusion methods.
Hence, we do not utilize this augmentation mechanism in our framework for applicability and generality.

\subsection{Comparisons with State-of-the-art Methods} \label{exp_compare_with_sotas}


\subsubsection{Evaluation on KITTI Dataset}


Table~\ref{kitti_results} presents the quantitative comparison with state-of-the-art 3D object detection methods on the KITTI test benchmark. As is shown, EPNet~\cite{huang2020epnet} and EPNet++ outperform our LiDAR-based baseline PointRCNN by a large margin, yielding an improvement of 3.64\% and 6.32\% on the moderate Cars, respectively. These results demonstrate the effectiveness of the proposed fusion mechanism. Besides, EPNet++ achieves better or comparable performance over the state-of-the-art methods on Cars and Pedestrians for all difficulty levels. For Cyclists, we find that their proportion of  instances on the training KITTI dataset is about~4.67\%, which is much smaller than that of Pedestrians~(12.87\%) and Cars~(82.46\%). In this case, constructing more instances with the GT Sampling strategy can effectively improve the detection performance for Cyclists. Hence, our method achieves relatively lower detection performance on Cyclists compared to a few methods~\cite{shi2020pv,yang2019std,yang20203dssd} using GT Sampling. However, EPNet++ achieves consistent performance improvement over the baseline PointRCNN on all difficulty levels for Cyclists, further demonstrating the effectiveness of our approach.

\subsubsection{Evaluation on JRDB Dataset}

In Table~\ref{jrdb_reults}, we compare our EPNet++ with the available published methods F-PointNet~\cite{qi2018frustum} and TANet~\cite{liu2020tanet} provided by the JRDB 3D detection benchmark. We also present our baseline configuration named EPNet~(LiDAR-only) and EPNet~\cite{huang2020epnet}.
F-PointNet achieves a lower detection performance than other detection methods. The main reason is that F-PointNet predicts 3D boxes from point cloud frustums, which are highly dependent on the performance of the 2D detector. Compared with the LiDAR-only method TANet~\cite{liu2020tanet}, our fusion-based methods EPNet++ and EPNet respectively obtain performance improvement with the mAP of 11.69\% and 8.65\%, which illustrates the superiority of our fusion manner. Besides, EPNet++ outperforms the EPNet and EPNet~(LiDAR-only) methods with the mAP of 3.04\%  and 7.38\%, respectively, which further verifies the effectiveness of our method.

\begin{table}[htbp]
\scriptsize
\centering
\caption{Quantitative comparisons on JRDB test set for 3D object detection. Ped. is short for Pedestrians. EPNet~(LiDAR-only) removes the image stream and LI-Fusion module of the original EPNet~\cite{huang2020epnet}. }
\begin{tabular}{l|c|c|c|c}
\hline
Method &Conference &Class &Modality & mAP \\
\hline
\hline
F-PointNet~\cite{qi2018frustum} &CVPR 2018 & Ped.  & P + I     &38.21   \\
TANet~\cite{liu2020tanet}       &AAAI 2020 & Ped.  & P         &54.94   \\
\hline
EPNet~(LiDAR-only)~\cite{huang2020epnet} &ECCV 2020 & Ped.  & P   &59.25   \\
EPNet~\cite{huang2020epnet} &ECCV 2020 & Ped.  & P + I     &63.59   \\
EPNet++~(Ours)              &- & Ped.  & P + I     &\textbf{66.63}  \\
\hline
\end{tabular}
\label{jrdb_reults}
\end{table}


\subsubsection{Evaluation on SUN-RGBD Dataset}

To further verify the effectiveness and generalization capability of our methods, we integrate the LI-Fusion module and the CB-fusion module into another two representative 3D object detectors, including LiDAR-based VoteNet~\cite{qi2019deep} and fusion-based ImVoteNet~\cite{qi2020imvotenet}. We implement our method based on the open-source mmdetection3d codebase\footnote{\label{mmdet3d}https://github.com/open-mmlab/mmdetection3d}. As is shown in Table~\ref{sunrgbd_results}, LI-Fusion and CB-Fusion modules bring an obvious improvement of 1.8\% and 2.4\% mAP, when using VoteNet as the backbone. ImVoteNet~\cite{qi2020imvotenet} boosts the detection performance of VoteNet~\cite{qi2019deep} via introducing geometric cues, semantic cues, and texture cues from the 2D camera images to point cloud features. Although effective, ImVoteNet~\cite{qi2020imvotenet} simply combines these image features with point cloud features in a concatenation manner at the end of the network, without exploring the complementarity between two modalities at a deep feature level. Adding our LI-Fusion and CB-fusion modules into ImVoteNet~\cite{qi2020imvotenet} leads to an  improvement of mAP 0.6\% and 1.3\%, respectively, which indicates that our fusion mechanism  produces more informative feature representations. Besides, CB-Fusion leads to consistent improvements over LI-Fusion, demonstrating the effectiveness of the bi-directional fusion strategy. It is worth noting that ImVoteNet~\cite{qi2020imvotenet} with CB-Fusion achieves a new state-of-the-art result on the SUN-RGBD dataset and outperforms all published point-based and fusion-based methods, which clearly illustrates the superiority of the CB-Fusion module.

\subsection{Ablation Studies} \label{exp_ablation_study}
\begin{figure}[t!]
\centering
  \includegraphics[width=0.95\linewidth]{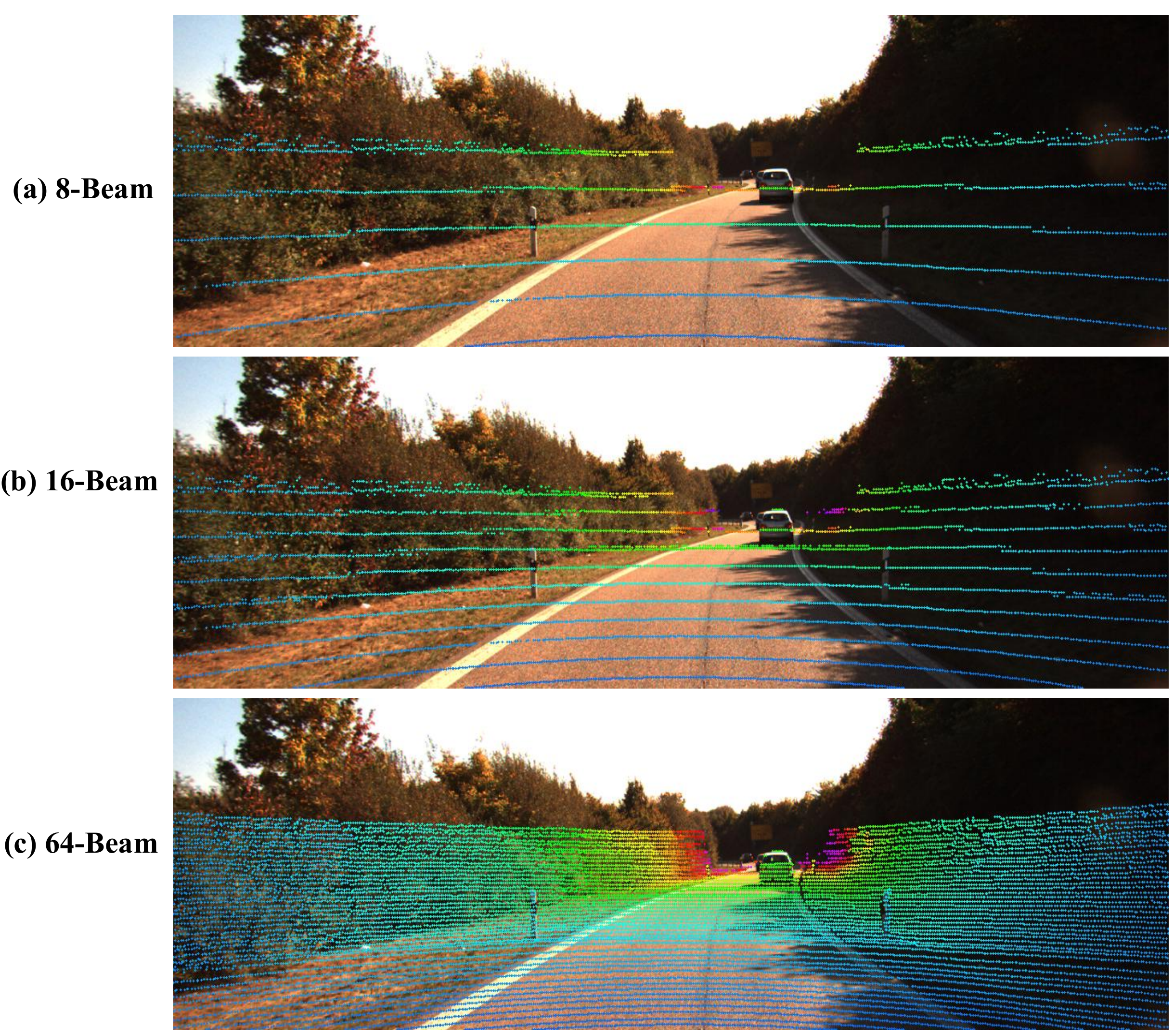}
\caption{Visualization of different beam LiDAR points projected on the image plane.}
\label{beam_vis}
\end{figure}

In this section, we conduct all the experiments on the KITTI validation dataset. \zliu{We provide results under different LiDAR beam settings~(64 beams, 16 beams and 8 beams) to investigate the generalization capability of our model to sparse scenes. To simulate real-world sparse scenes,
we generate 8-beam and 16-beam LiDAR points following the open-source code \footnote{\label{pseudo_lidar}https://github.com/mileyan/Pseudo\_Lidar\_V2} in Pseudo-Lidar-V2~\cite{you2020pseudo}. Fig.~\ref{beam_vis} illustrates a visualization result of generated 16-beam and 8-beam LiDAR points using the algorithm above. In the following, we take the 64-beam LiDAR points as input, unless otherwise stated.} Since CE loss is well verified in our preliminary work EPNet~\cite{huang2020epnet}, we add it to the optimization goal by default and do not discuss the effect of CE loss here. For more details about CE loss, please refer to EPNet~\cite{huang2020epnet}.

\subsubsection{Comparisons under Different Point Cloud Densities}

\begin{table*}[htbp]
\scriptsize
\centering
\caption{Comparisons with state-of-the-art approaches under different point cloud densities\zliu{~(8-beam, 16-beam and 64-beam)} on the KITTI validation dataset. P and I are short for point cloud and camera image.} 
\begin{tabular}{l|c|c|c|c|c|c|c|c|c|c|c|c}
\hline
\multirow{2}{*}{Method} 
& \multirow{2}{*}{GT Sampling}
& \multicolumn{1}{c|}{\multirow{2}{*}{\shortstack{Input}}} 
& \multicolumn{1}{c|}{\multirow{2}{*}{\shortstack{{LiDAR}}}} 
& \multicolumn{3}{c|}{Cars} 
& \multicolumn{3}{c|}{Pedestrians} 
& \multicolumn{3}{c}{Cyclists} \\
\cline{5-13}
& \multicolumn{1}{c|}{} & & & Easy & Mod. & Hard  & Easy & Mod. & Hard & Easy & Mod. & Hard\\
\hline
\hline
\multirow{3}{*}{Voxel RCNN~\cite{deng2021voxel}} 
& \multirow{6}{*}{\checkmark} 
& \multirow{3}{*}{P}
& \zliu{8-beam} &{69.10} &{51.69} &{46.44} &{40.71} &{35.21} &{30.48} &{38.48} &{22.96} &{21.66} \\ 
& & &\zliu{16-beam} &{86.55} &{68.70} &{63.93} &{57.07} & {51.26} &{46.28} &{66.87} &{43.34} &{40.46} \\  
& & &\zliu{64-beam} &\textbf{92.57} &84.49 &82.35 &68.63 &60.95 &55.67 &93.16 &\textbf{76.52} &\textbf{72.57} \\
\cline{1-1} \cline{3-13}
\multirow{3}{*}{PV RCNN~\cite{shi2020pv}} 
& & \multirow{3}{*}{P} 
&\zliu{8-beam} &{68.30} &{49.33} &{44.03} &{42.30} &{37.13} &{32.40} &{40.29} &\textbf{23.61} &\textbf{22.64} \\  
& & &\zliu{16-beam} &{85.60} &{66.96} &{62.78} &{59.92} &{52.79} &{48.27} &\textbf{68.64} &\textbf{45.67} &\textbf{42.94} \\ 
& & &\zliu{64-beam} &92.35 &\textbf{84.50} &\textbf{82.47} &66.39 &60.34 &55.70 &\textbf{93.71} &73.99 &69.53 \\
\hline
\hline
\multirow{3}{*}{Voxel RCNN~\cite{deng2021voxel}} & \multirow{21}{*}{$\times$} 
& \multirow{3}{*}{P} &\zliu{8-beam} &{65.30} &{48.03} &{43.18} &{42.03} &{35.58} &{31.74} &{25.60} &{14.39} &{13.24} \\  
& & &\zliu{16-beam}  &{85.18} &{66.35} &{61.54} &{58.64} &{53.34} &{47.40} &{55.03} &{33.83} &{31.93} \\  
& & &\zliu{64-beam} &92.04 &82.64 &80.23 &70.94 &64.55 &58.35 &85.12 &61.53 &58.93 \\  
\cline{1-1} \cline{3-13}
\multirow{3}{*}{PV RCNN~\cite{shi2020pv}} & 
&\multirow{3}{*}{P}  &\zliu{8-beam} &{62.55} &{46.77} &{41.08} & {34.53} &{29.38} &{26.11} &{22.95} &{12.80} &{12.52} \\   
& & &\zliu{16-beam} &{83.73} &{65.36} &{60.80} &{54.32} &{46.74} &{42.68} &{47.37} &{31.71} &{29.97} \\   
& & &\zliu{64-beam} &91.85 &82.68 &80.20 &67.49 &62.21 &57.91 &80.04 &58.72 &56.36 \\ 
\cline{1-1} \cline{3-13}
\multirow{3}{*}{CLOCs~\cite{pang2020clocs}} & 
& \multirow{3}{*}{P+I} 
&\zliu{8-beam} &{67.35} &{52.98} &{46.02} &- &- &- &- &- &- \\ 
& & & \zliu{16-beam}  &{85.41} &{67.26} &{62.02} &- &- &- &- &- &- \\ 
& & & \zliu{64-beam} &{92.48} &{82.79} &{77.71} &- &- &- &- &- &-  \\ 

\cline{1-1} \cline{3-13}
\multirow{3}{*}{3D-CVF~\cite{yoo20203d}} 
& & \multirow{3}{*}{P+I}  &\zliu{8-beam} &{61.95} &{46.67} &{40.50}  &- &- &- &- &- &- \\ 
& & & \zliu{16-beam}  &{81.82} &{63.12} &{57.19} &- &- &- &- &- &- \\  
& & & \zliu{64-beam} &91.97 &82.87 &80.36 &- &- &- &- &- &-  \\ 
\cline{1-1} \cline{3-13}
\multirow{3}{*}{EPNet~(LiDAR)} 
&  &\multirow{3}{*}{P} &\zliu{8-beam} &{64.55} &{47.10} &{42.32} &{41.81} &{36.16} &{31.15} &{33.07} &{18.88} &{17.99} \\ 
& & &\zliu{16-beam}  &{84.47} &{63.30} &{59.59} &{58.47} &{52.56} &{45.73} &{55.94} &{33.80} &{31.59} \\ 
& & &\zliu{64-beam} &90.87	&81.15  &79.59  &70.26	&61.30	&54.16 	&84.49  &61.50  &57.86 \\ 

\cline{1-1} \cline{3-13}
\multirow{3}{*}{EPNet~\cite{huang2020epnet}}
& & \multirow{3}{*}{P+I} &\zliu{8-beam} &{69.11} &{53.90} &{49.06} &{46.65} &{40.62} &{35.36} &{35.41} &{20.73} &{19.13} \\ 
& & &\zliu{16-beam} &{85.11} &{67.72} &{63.32}  & {61.08} &{54.16} &{47.71} &{58.01} &{34.53} &{32.47} \\   
& & &\zliu{64-beam} &92.28 &82.59  &80.14  &72.56  &63.11  &56.32 	&84.88  &62.43  &58.65 \\ 
\cline{1-1} \cline{3-13}
\multirow{3}{*}{EPNet++~(Ours)} 
& & \multirow{3}{*}{P+I} 
& \zliu{8-beam} &\textbf{69.20} &\textbf{55.11} &\textbf{50.29} &\textbf{51.31} &\textbf{45.79} &\textbf{39.78} &\textbf{41.34} &{23.31} &{21.68} \\ 
& & &\zliu{16-beam} &\textbf{87.08} &\textbf{69.12} &\textbf{64.90} &\textbf{66.77} &\textbf{59.86} &\textbf{53.07} &{61.37} &{36.93} &{34.95} \\ 
& & &\zliu{64-beam} &92.51 &83.17 &82.27 &\textbf{73.77} &\textbf{65.42} &\textbf{59.13} &86.23 &63.82 &60.02 \\ 
\hline
\end{tabular}
\label{sparse_cmp_results}
\end{table*}

\begin{table*}[htbp]
\scriptsize
\centering
\caption{Ablation experiments on the effects of different components of the proposed EPNet++ on \zliu{16-beam and 64-beam LiDAR points}. LI~(CB) wo att represents the LI-Fusion~(CB-Fusion) without attention mechanism. MC is the MC loss.} 
\begin{tabular}{l|ccccc|c|c|c|c|c|c|c|c|c}
\hline
\multicolumn{1}{c|}{\multirow{2}{*}{LIDAR}} 
&\multicolumn{1}{c}{\multirow{2}{*}{\zliu{LI wo Att}}}
& \multicolumn{1}{c}{\multirow{2}{*}{{LI}}}  
&\multicolumn{1}{c}{\multirow{2}{*}{\zliu{CB wo Att}}}
&\multicolumn{1}{c}{\multirow{2}{*}{{CB}}}
& \multicolumn{1}{c|}{\multirow{2}{*}{{MC}}} 
& \multicolumn{3}{c|}{{Cars}} 
& \multicolumn{3}{c|}{{Pedestrians}}
& \multicolumn{3}{c}{{Cyclists}}  \\
\cline{7-15}
& \multicolumn{3}{c}{} & & & {Easy} & {Moderate} & {Hard} & {Easy} & {Moderate} & {Hard}  & {Easy} & {Moderate} & {Hard}  \\
\hline
\zliu{16-beam}  &- &- &-	&-  &-  &84.47 &63.30 &59.59 &58.47 &52.56 &45.73 &55.94 &33.80 &31.59  \\ 
\zliu{16-beam}   &\checkmark &- & -	&- &-  &84.60	&67.63  &63.20   & 59.47 &53.26  &46.87 &57.45 & 33.93 & 31.96  \\ 
\zliu{16-beam}   &- &\checkmark &-  &-  &-  &85.11 &67.72 &63.32  & 61.08 &54.16 &47.71 &58.01 &34.53 &32.47 \\  
\zliu{16-beam}   &- &-  & \checkmark	&- &-  &86.10	&68.86  &63.84   & 61.86 &54.24	& 47.56  &58.12 &35.35 &32.99  \\ 
\zliu{16-beam}   &- & -	&- &\checkmark  &-    &85.69	&68.83  &64.22   & 65.73 &57.69	&51.13  &58.72 &37.25  &34.47  \\
\zliu{16-beam}   &- & -	&- &- &\checkmark      &84.50	&63.69  &60.22   & 61.95 &55.90	&49.02  &60.71 &35.96  &33.61  \\
\zliu{16-beam}   & - &- &- &\checkmark &\checkmark  &\textbf{87.08} &\textbf{69.12} &\textbf{64.90} &\textbf{66.77} &\textbf{59.86} &\textbf{53.07} &\textbf{61.37} &\textbf{36.93} &\textbf{34.95} \\
\hline
\zliu{64-beam} &- &- &-	&-  &- &90.99	&81.72  &79.75  &70.26	&61.30	&54.16 	&84.49  &61.50  &57.86 \\ 
\zliu{64-beam}  &\checkmark & -	&- &- &-  &91.92	&82.42  &80.16   & 71.02 &61.56 & 54.49  &84.79 &62.08 &58.24  \\  
\zliu{64-beam}  &- & \checkmark	&- &- &-  &92.28 &82.59  &80.14  &72.56  &63.11  &56.32 	&84.88  &62.43  &58.65 \\
\zliu{64-beam}  &- &- & \checkmark	&-  &-  &92.45	&82.73  &80.44  & 72.49 &61.96	&54.52  &84.69 &62.16  &58.19  \\ 
\zliu{64-beam}  &- & -	&-  &\checkmark &-  &92.54	&82.93  &80.56   & 73.08 &64.03 &57.65  &85.57 &62.92  &58.70  \\ 
\zliu{64-beam}  &- & -	&- &-  &\checkmark    &91.69	&82.12  &80.19   & 70.71 &62.76 &55.74  &85.46 &62.34 &58.25  \\ 
\zliu{64-beam} &-   &-  &- & \checkmark &\checkmark  &\textbf{92.51} &\textbf{83.17} &\textbf{82.27} &\textbf{73.77} &\textbf{65.42} &\textbf{59.13} &\textbf{86.23} &\textbf{63.82} &\textbf{60.02} \\
\hline
\end{tabular}
\label{Ablation_Experiments}
\end{table*}

To comprehensively evaluate the detection performance under dense and sparse scenes, we compare our method with several state-of-the-art detectors under three different LiDAR beam settings. Specifically, we select two LiDAR-based methods~(Voxel RCNN~\cite{deng2021voxel}\footnote{\label{OpenPCDet}https://github.com/open-mmlab/OpenPCDet}, PV-RCNN~\cite{shi2020pv}\textsuperscript{\ref{OpenPCDet}}), and two fusion-based methods~( CLOCs~\cite{pang2020clocs}\footnote{\label{CLOCs}https://github.com/pangsu0613/CLOCs}, 3D-CVF~\cite{yoo20203d}\footnote{\label{3dcvf}https://github.com/rasd3/3D-CVF}). Since the GT Sampling data augmentation usually effectively boosts the performance of LiDAR-based detectors while not applicable for fusion-based methods, we provide the results of LiDAR-based detectors with and without GT Sampling for fair comparison in Table~\ref{sparse_cmp_results}.

At the top of Table~\ref{sparse_cmp_results}, we first compare EPNet++ with state-of-the-art LiDAR-based methods which leverage the powerful GT Sampling augmentation. Although EPNet++ produces worse performance on Cars and Cyclists in dense 64-beam LiDAR setting, our method demonstrates clear superiority on Cars for sparse 8-beam LiDAR, outperforming Voxel R-CNN and PV-RCNN on the moderate difficulty level by 3.42\% and 5.78\% in terms of mAP. Besides, the performance gap on the Cyclists is also reduced. When GT Sampling is not employed, EPNet++ consistently outperforms previous methods for all the different settings of categories, difficulty levels and the number of beams, as is shown at the bottom of Table~\ref{sparse_cmp_results}. Specifically, for sparse scenes with 8-beam LiDAR points, EPNet++ yields reliable improvements with mAP of 2.13\% and 8.44\% over the fusion-based methods CLOCs and 3D-CVF on Cars of moderate difficulty level, respectively. 
For more challenging Pedestrians in the case of 8-beam LiDAR points as input, EPNet++ attains the obvious gains with the mAP of 10.21\%, 16.41\% and 9.63\% over these LiDAR-based methods Voxel-RCNN, PV-RCNN and EPNet~(LiDAR-only) without GT Sampling on the moderate difficulty level. 
Moreover, EPNet++ achieves consistent and remarkable improvements over EPNet, which reveals the effectiveness of our new design for multi-modal 3D object detection tasks.

\subsubsection{Ablation Studies on Different Components}

In this part, we validate the effectiveness of core components of our EPNet++, including LI-Fusion with/without attention, CB-Fusion with/without attention and MC loss. We build our baseline by removing the image stream and these components from EPNet++ for both 64-beam and 16-beam LiDAR points. As is shown in Table~\ref{Ablation_Experiments}, CB-Fusion~(or LI-Fusion) without attention brings general and consistent improvements over the baseline on Cars, Pedestrians and Cyclists under all the difficulty settings on different beam LiDAR points, which demonstrates the effectiveness of the cascade bi-directional feature enhancement pathway. Then, \zliu{introducing attention operations into CB-Fusion~(or LI-Fusion) brings a remarkable gain with mAP of 3.45\%~(or 0.90\%) on Pedestrians under the case of 16-beam LiDAR points, which reveals the importance of enhancing reliable features and suppressing harmful ones by computing the relevance of image and point features.} Moreover, the CB-Fusion with attention module yields an improvement of 1.11\%, 3.53\% and 2.72\% over the LI-Fusion module in terms of mAP on Cars, Pedestrians and Cyclists for the moderate difficulty level under the case of 16-beam LiDAR points, which further illustrates the superiority of adopting feature interaction in a bi-directional manner. Therefore, \zliu{CB-Fusion is the most indispensable component in EPNet++ for promising performance especially under the sparse setting.}
Beyond the feature fusion, MC loss leads to consistent improvements across all the categories and difficulty levels on different beam LiDAR points. The reason is that MC loss effectively aligns the confidence scores from two streams and the aligned score is a better criterion for evaluating the quality of 3D proposals in the RPN stage. Combining CB-Fusion with MC loss, EPNet++ achieves significant gains with mAP of 5.82\%, 7.30\%, 3.13\% on Cars, Pedestrians and Cyclists at the moderate difficulty level over the baseline for 16-beam LiDAR points. \zliu{It is worth mentioning that MC loss is easy to implement without introducing extra computation cost in the inference stage, which is friendly to a more computing-sensitive application.}

\subsubsection{Different Fusion Paradigms}

\begin{figure}[htbp]
\centering
  \includegraphics[width=0.9\linewidth]{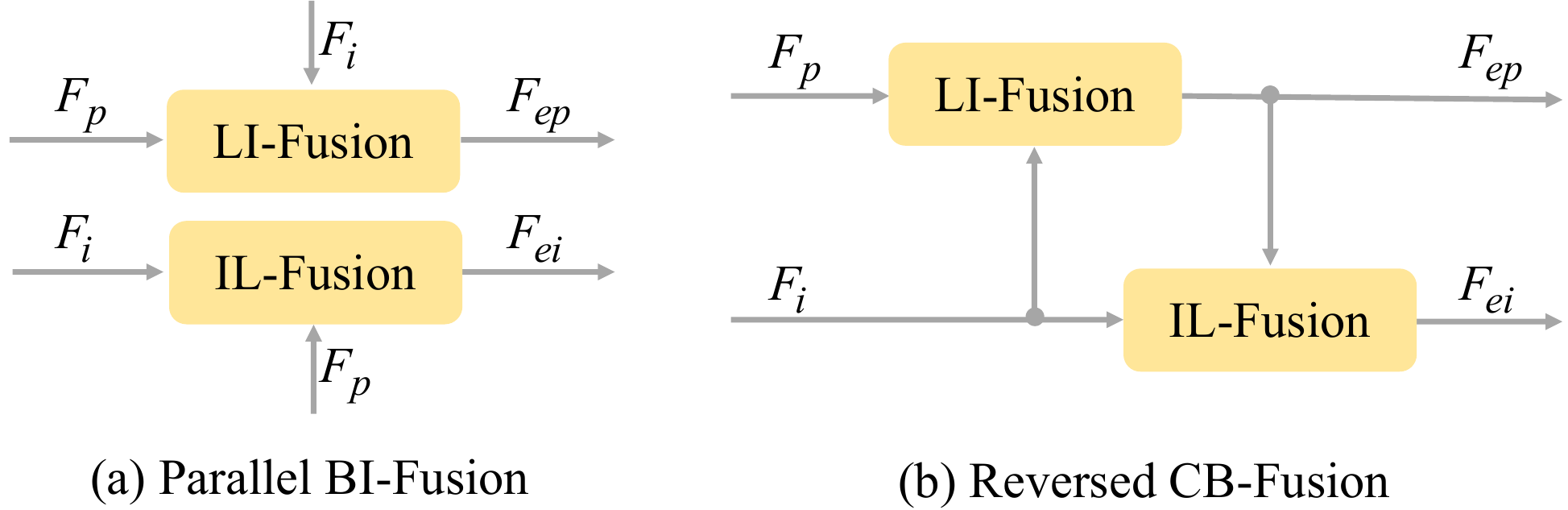}
\caption{The structure of two alternative fusion mechanisms: the \emph{parallel bi-directional interaction fusion}~(Parallel BI-Fusion) in (a) and the reversed CB-Fusion in (b).}
\label{vis_diff_fusion}
\end{figure}

\begin{table}[htbp]
\scriptsize
\centering
\caption{Comparisons with different bi-directional fusion mechanisms. For each category, we report the average value of mAP under three difficulty levels \zliu{ for 64-beam LiDAR points. } }
\begin{tabular}{c|c|c|c|c}
\hline
\zliublack{Methods} & \zliublack{Cars} & \zliublack{Pedestrians} & \zliublack{Cyclists} & \zliublack{mAP}  \\
\hline
EPNet++ w/o CB-Fusion &84.67 &63.07  &68.68  & 72.14   \\ 
\zliublack{Parallel BI-Fusion}   &85.49	&64.47  &69.78   & 73.25   \\ %
\zliublack{Reversed CB-Fusion}   &85.41	&65.07  &69.69  & 73.39   \\ 
\zliublack{CB-Fusion} 	&\textbf{85.98} 	&\textbf{66.11}  &\textbf{70.02}   &\textbf{74.04} \\
\hline
\end{tabular}
\label{exp_diff_fusion_1}
\end{table}

\noindent\textbf{Different Bi-Directional Fusion Mechanisms.} In Table~\ref{exp_diff_fusion_1}, we provide two alternative CB-Fusion methods, namely, a parallel BI-Fusion shown in Fig.~\ref{vis_diff_fusion}(a) and a reversed CB-Fusion shown in Fig.~\ref{vis_diff_fusion}(b). EPNet++ without the CB-Fusion module is regarded as our baseline model. The core difference between parallel BI-Fusion and (reversed) CB-Fusion is whether to use the enhanced features for the attention operation. Parallel BI-Fusion conducts two attention operations parallelly using the original image/point features. However, CB-Fusion and reversed CB-Fusion work in a cascaded manner, and the second attention operation takes as input the enhanced feature~(i.e., the output of the first attention operation). \zliu{Intuitively, the enhanced feature, which has already combined image semantic information with geometric information, will be more beneficial.} Experimental results in Table~\ref{exp_diff_fusion_1} demonstrate the clear superiority of using the enhanced feature. CB-Fusion~(resp. reversed CB-Fusion) outperforms parallel BI-Fusion by 0.79\%~(resp. 0.14\%) in terms of mAP. Reversed CB-Fusion differs from CB-Fusion in the fusion order. \zliu{Since the geometric stream~(point feature) is finally applied for predicting 3d boxes, CB-Fusion is more reasonable and leads to  better performance than reversed CB-Fusion~(74.04\% v.s. 73.39\% in terms of mAP).}

\begin{table}[t!]
\scriptsize
\centering
\caption{\zliu{Comparisons with different fusion methods. For each category, we report the average value of mAP under three difficulty levels  for 64-beam LiDAR points.}} \zliuall
\begin{tabular}{c|c|c|c|c}
\hline
{Methods} & {Cars} & {Pedestrians} & {Cyclists} & {mAP}  \\
\hline
LiDAR-only EPNet &83.87 &61.91  &67.95 & 71.24 \\
PointPainting~\cite{vora2020pointpainting} &84.48	&63.37  &68.58  & 72.14   \\ 
Pointaugmenting~\cite{wang2021pointaugmenting} &84.13	&63.60  &69.12  & 72.28   \\
DenseFusion~\cite{wang2019densefusion} &84.94 &63.35  &68.80  & 72.36   \\  %
Transformer-based CB-Fusion 	&84.52 &64.54  &68.43   & 72.50  \\ 
CB-Fusion 	&\textbf{85.34} 	&64.92  &69.07   & 73.11  \\ %
Pointaugmenting+CB-Fusion 	&84.56 	&\textbf{66.30}  &\textbf{71.95}   & \textbf{74.27}  \\ %
\hline
\end{tabular}
\label{exp_diff_fusion_2}
\end{table}

\noindent\zliu{\textbf{Comparisons with different fusion methods.} In Table~\ref{exp_diff_fusion_2},  we provide the quantitative comparisons for CB-Fusion, transformer-based CB-Fusion, PointPainting-based methods~\cite{vora2020pointpainting,wang2021pointaugmenting} and DenseFusion~\cite{wang2019densefusion}.  The baseline model is LiDAR-only EPNet for a fair comparison. First, we integrate the popular
PointPainting-based fusion methods PointPainting~\cite{vora2020pointpainting} and Pointaugmenting~\cite{wang2021pointaugmenting} into our baseline. More concretely, we first use Mask-RCNN~\cite{he2017mask} to train a strong segmentation network.  For PointPainting~\cite{vora2020pointpainting}, we concatenate the output segmentation scores and input LiDAR points to achieve multi-modal fusion. For Pointaugmenting~\cite{wang2021pointaugmenting}, the input LiDAR points are decorated with the corresponding image features from the first output layer after the backbone of ResNet-50~\cite{he2016deep} from Mask-RCNN in a concatenation manner. As shown in Table~\ref{exp_diff_fusion_2}, our CB-Fusion outperforms these methods with input fusion on average. Besides, integrating CB-Fusion into the input-feature fusion PointAugmenting produces a performance gain of 1.99\% mAP on average, which further demonstrates the effectiveness of this bidirectional feature interaction between LiDAR points and images. Then, we conduct the pixel-wise dense fusion similar to DenseFusion\cite{wang2019densefusion} at object levels in the RCNN stage of EPNet++ by combining 3D RoI features and the corresponding projected 2D boxes feature in a concatenation manner. As shown in Table~\ref{exp_diff_fusion_2}, DenseFusion brings a reliable improvement of 1.12\% mAP~(72.36 v.s. 71.24) on average over the baseline, but is still inferior to CB-Fusion~(73.11\%). Finally, we provide the transformer-based CB-Fusion through a global cross-attention operation~(refer to Section~\ref{trans_based_cb} for details) instead of attention operation in LI-Fusion and IL-Fusion, which even produces a slight performance drop. This reveals effectively utilizing the one-to-one correspondence between two modalities is crucial in computing attention weights.}

\subsubsection{Effect of the Attention Gate Weights}

\begin{figure}[htbp]
\centering
  \includegraphics[width=0.98\linewidth]{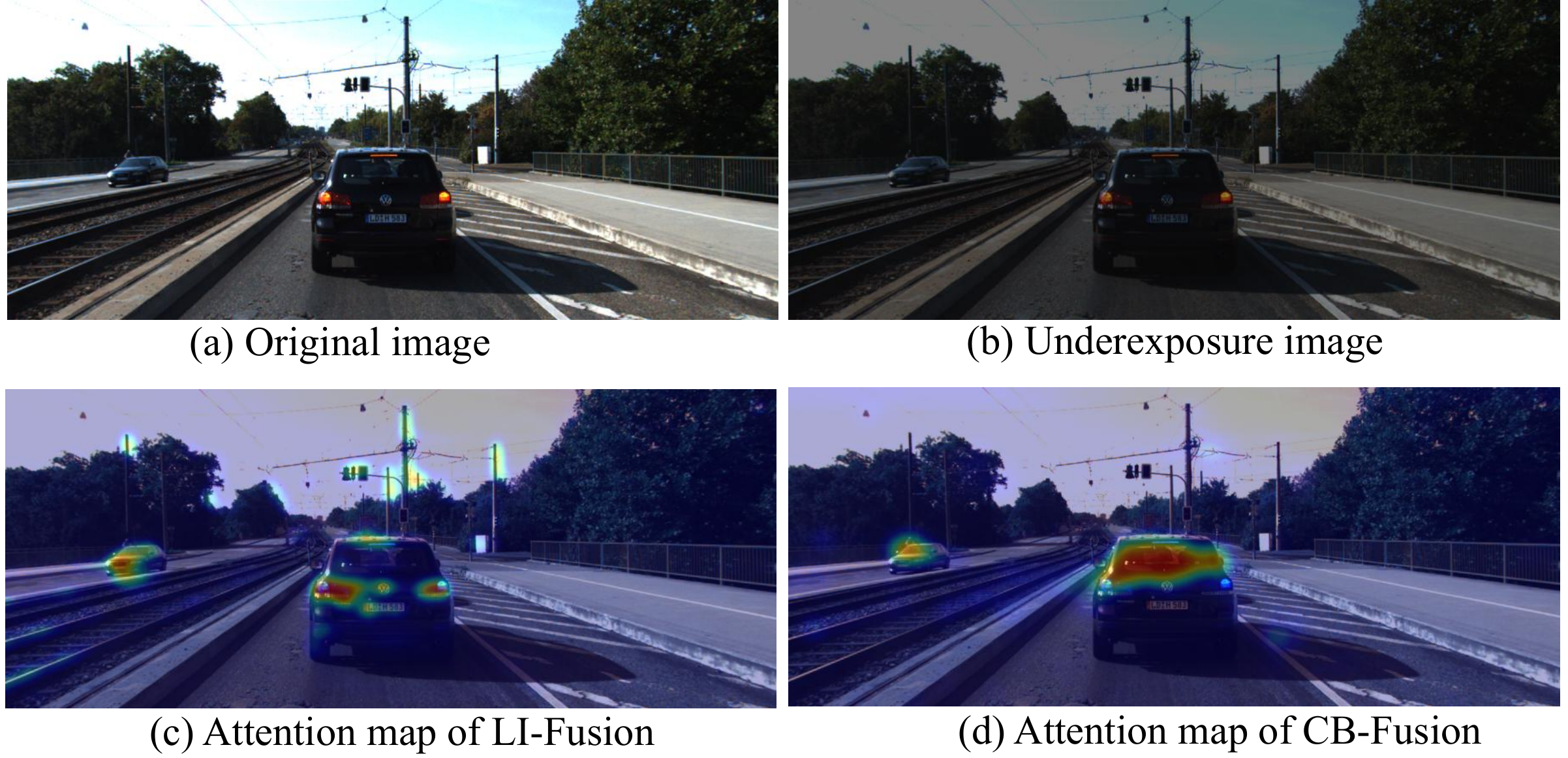}
\caption{Visualization of the attention map for LI-Fusion and CB-Fusion, respectively.}
\label{grad_cam}
\end{figure}

\begin{table}[htbp]
\scriptsize
\centering
\caption{Quantitative analysis of the effect of I2P attention gate $w_{i2p}$ and P2I attention gate  $w_{p2i}$. For each category, we report the average value of mAP under three difficulty levels. }
\begin{tabular}{ccc|c|c|c|c}
\hline
Method & $w_{i2p}$ & $w_{p2i}$ &Cars & Pedestrians & Cyclist & mAP \\
\hline
Baseline &- & - & {83.00} &{58.28} & {62.39} & {67.89} \\
LI-Fusion  &- & - & {82.63} & {57.73} & {62.95} & {67.77} \\ 
LI-Fusion  &\checkmark & - & {83.03} & {59.19} & {66.66} & {69.63} \\
CB-Fusion  &- & - & {83.50} & {57.17} & {65.92} & {68.86} \\
CB-Fusion  &\checkmark & \checkmark & \textbf{83.83} & \textbf{60.98} & \textbf{67.89}  & \textbf{70.90} \\
\hline
\end{tabular}
\label{robustness_exp}
\end{table}

In a real scene, the camera image usually suffers from underexposure and overexposure due to the change of illumination. Furthermore, the point cloud collected by a LiDAR sensor is easily disturbed by severe weather~(\textit{e.g.}, fog, rain) conditions, resulting in noise points. It poses challenges for 3D detectors to produce accurate predictions under these scenes. We argue that the attention gate weights employed in our fusion module can effectively alleviate the issue of noisy images and point clouds. To investigate the effect of the attention gate weights, we simulate the real noising scenes by changing the illumination of the camera image and adding inference points into the raw LiDAR point cloud. Specifically, for each image in the KITTI dataset, we simulate the illumination variance through the transformation $y=a*x+b$, where $x$ and $y$ denote the original and transformed RGB value for a pixel. The coefficient $a$ is randomly sampled from a uniform distribution in the range of [0.5, 1.5]. The offset $b$ is set as a fixed value of 5. For a raw point cloud on the KITTI dataset, we generate 100 noise points around each GT 3D object following TANet~\cite{liu2020tanet}.

The experimental results are presented in Table~\ref{robustness_exp}. We first present a LiDAR-based baseline result by removing the image stream of EPNet, which produces 67.89\% mAP averaged across Cars, Pedestrians and Cyclists. Then we introduce the LI-Fusion module without I2P attention gate weight into the baseline, leading to slight performance degradation of 0.12\% on average, which indicates that combining noisy sensor data is harmful to 3D object detection. In contrast, with the guidance of the I2P attention gate weight $w_{i2p}$, the LI-Fusion module yields a gain of 1.74\% mAP over the baseline. Similarly, integrating the attention gate weights $w_{i2p}$ and $w_{p2i}$ into the CB-Fusion module brings a remarkable performance improvement of 2.04\% mAP, which further demonstrates the importance of the attention gate weights for dealing with the challenging scenes especially when the sensor data is disturbed. 
\zliu{Besides, as shown in Fig.~\ref{grad_cam}, we visualize the attention map of the block from the image stream in the case of illumination interference~(\textit{e.g.}, $a=0.5$ and $b=5$ for underexposure image in Fig.~\ref{grad_cam}~(b)) through Grad-CAM~\cite{selvaraju2017grad}.  The attention maps from LI-Fusion  in Fig.~\ref{grad_cam}~(c) and CB-Fusion in Fig.~\ref{grad_cam}~(d) mainly focus on the region of foreground objects. Besides, compared with LI-Fusion, CB-Fusion pays more attention to  more discriminative foreground regions and ignores irrelevant objects~(\textit{e.g.}, traffic signs).
}


\subsubsection{Effect of MC Loss on Score Alignment}

\begin{table}[htbp]
\scriptsize
\centering
\caption{Statistical analysis of the L1 error of confidence scores from two modalities in terms of mean and variance.\textit{Var.} is short for Variance.}
\begin{tabular}{c|c|c|c|c|c|c|c}
\hline
\multirow{2}{*}{\shortstack{{LiDAR}}} &\multirow{2}{*}{MC Loss} &\multicolumn{2}{c|}{Cars} &\multicolumn{2}{c|}{Pedestrians} &\multicolumn{2}{c}{Cyclists} \\
\cline{3-8}
& &Mean &Var. &Mean &Var. &Mean &Var.  \\
\hline
\multirow{2}{*}{\zliu{16-beam}}
&{-}       &{4.70} &{9.50} &{4.19} &{14.76} &{4.50} &{22.12} \\ 
&{\checkmark}  &\textbf{4.49} &\textbf{0.62} &\textbf{3.24} &\textbf{0.95} &\textbf{3.27} &\textbf{0.79} \\ %
\hline
\multirow{2}{*}{64-beam}
&-       &4.51 &15.48 &3.41 &6.84 &3.27 &28.33 \\ 
&\checkmark  &\textbf{1.52} &\textbf{0.51}  &\textbf{0.64} &\textbf{0.14} &\textbf{1.63} &\textbf{1.13} \\ %
\hline
\end{tabular}
\label{satt_mc_results}
\end{table}

To analyze the effect of MC loss on promoting the consistency of confidence scores from two modalities, we compute their L1 error in terms of the mean and variance of confidence scores with or without  MC loss. Table~\ref{satt_mc_results} presents the L1 error for three categories and different beam LiDAR points. It is obvious that MC loss narrows the difference in confidence scores predicted by two streams, which indicates that MC loss is beneficial for generating more stable and reliable scores.

\begin{figure*}[htbp]
\centering
  \includegraphics[width=0.96\linewidth]{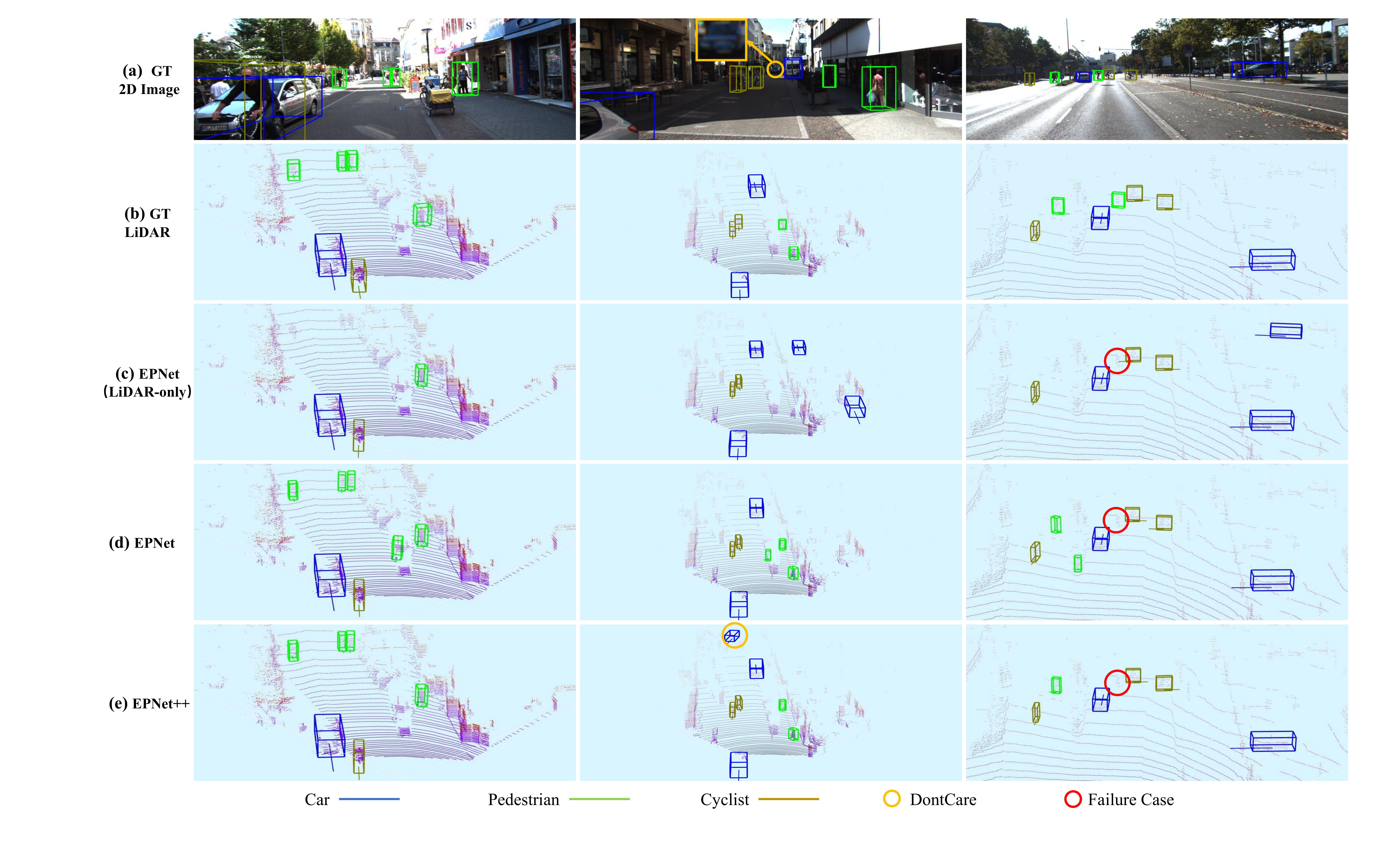}
\caption{The visualization on the KITTI dataset. We localize the objects of Cars, Pedestrians and Cyclists with the blue, green and brown 3D bounding boxes, respectively. Besides, we highlight the DontCare object and the failure case~(\textit{e.g.}, false negative) with the yellow and red circle box.}
\label{vis_kitti_results}
\end{figure*}

\subsubsection{Influence of $\lambda_1$ and $\lambda_2$} \label{exp_lambda}

\begin{table}[htbp]
\scriptsize
\centering
\caption{Quantitative analysis of the hyperparameter $\lambda_1$ and $\lambda_2$. For each category, we report the average value of mAP under three difficulty levels.}
\begin{tabular}{ccc||c|c|c|c}
\hline
{LiDAR} & $\lambda_1$  & $\lambda_2$ & Cars & Pedestrians & Cyclists & mAP    \\
\hline
\zliunew{16-beam}  &\zliunew{0.0}  &\zliunew{0.0}  &\zliunew{73.16}   &\zliunew{58.00}   &\zliunew{44.43}   &\zliunew{58.53}   \\ 
{16-beam}  &0.0  &1.0  &{72.41}   & {58.69}  &\textbf{45.35}  &{58.82}  \\  
{16-beam}  &0.5  &0.5  &{73.70}   & \textbf{59.90}  &{44.42} &\textbf{59.34}  \\ 
{16-beam}  &1.0  &0.0  &\textbf{74.40}   & {56.24}  &{44.30}  &{58.31}  \\ 
\hline
\zliunew{64-beam}  &\zliunew{0.0}  &\zliunew{0.0}  &\zliunew{85.11}   &\zliunew{65.16}  &\zliunew{69.19}   &\zliunew{73.15}   \\  
64-beam  &0.0  &1.0  &85.26   & 65.75 &\textbf{70.57}  &73.86  \\ 
64-beam  &0.5  &0.5  &\textbf{85.98}   & \textbf{66.11}  &{70.02} &\textbf{74.04}  \\ 
64-beam  &1.0  &0.0  &85.29   & 65.00 &66.74  &72.34  \\  
\hline
\end{tabular}
\label{lambda_exp}
\end{table}

As shown in Tab.~\ref{lambda_exp}, we conduct extensive experiments by varying the values of $\lambda_1$ and $\lambda_2$ in MC loss to analyze their impact on the detection performance under different densities of points.  \zliunew{Note that the average confidence score of the image and geometric stream is used as the default threshold in the nms procedure. Setting both $\lambda_1$ and $\lambda_2$ to 0 means the MC loss is not employed, which can serve as the baseline. MC loss outperforms the baseline by 0.81\% and 0.89\% in terms of mAP under 16-beam and 64-beam LiDAR setting respectively, demonstrating the effectiveness of the MC loss in selecting high-quality boxes.} Besides, for sparse scenes with 16-beam LiDAR points, setting $\lambda_1=0.0$ and $\lambda_2=1.0$ produces better detection results on Pedestrians and Cyclists than the setting of $\lambda_1=1.0$ and $\lambda_2=0.0$. It indicates that the confidence scores predicted by the geometric stream are less reliable especifically for these objects with fewer collected points in sparse scenes, and thus more attention should be paid to the scores generated by the image stream.

\subsubsection{Scalability of CB-Fusion}\label{trans_based_cb}

\begin{table}[htbp] \zliuall
\scriptsize
\centering
\caption{Result of Cars for three difficulty levels on KITTI validation. \zliunew{$^\dag$ represents the results with GT sampling strategy~\cite{yan2018second} enabled.}}
\begin{tabular}{c|c|c|c|c}
\hline
3D Backbone & Methods  & Easy & Moderate & Hard   \\
\hline
\multirow{3}{*}{Transformer-Based} 
 & Pointformer~\cite{pan20213d}   &91.06   & 81.20   &79.74   \\  
 & +CB-Fusion   &\textbf{92.52}  &\textbf{82.74}  & \textbf{80.39}       \\
 & +Trans. CB-Fusion &91.99 & 81.47 & 79.68 \\
\hline
\hline
\multirow{4}{*}{Voxel-Based} 
& SECOND~\cite{yan2018second}   &88.72   & 78.13   &74.03  \\  
& +CB-Fusion   &\textbf{89.13}  &\textbf{79.47}  & \textbf{74.88}     \\ 
\cline{2-5}
& Voxel-RCNN~\cite{deng2021voxel}   &92.04  &82.64  & 80.23     \\ 
& +CB-Fusion   &\textbf{92.64}  &\textbf{83.51}  & \textbf{80.77}      \\
\cline{2-5}
& Voxel-RCNN~\cite{deng2021voxel}$^\dag$  & \zliunew{92.57} & \zliunew{84.49} & \zliunew{82.35}     \\ 
& +CB-Fusion   & \zliunew{\textbf{92.89}}  & \zliunew{\textbf{85.20}}  &  \zliunew{\textbf{83.06}}      \\
 \hline 
\end{tabular}
\label{voxel_kitti}
\end{table}


\noindent\zliu{\textbf{Transformer-based CB-Fusion.} Recent approaches~\cite{pan20213d,zhao2021point,guo2021pct,han2021point} have shown the effectiveness of transformers for dealing with point clouds. In this section, we take Pointformer~\cite{pan20213d} as the backbone network to investigate the application of our CB-Fusion in transformer-based architecture. As shown in the top of Table~\ref{voxel_kitti}, integrating CB-Fusion into Pointformer leads to a performance gain with mAP of 1.54\% on the moderate Cars of the KITTI validation set, which demonstrates the effectiveness of CB-Fusion on the transformer-based 3D backbone. Moreover, motivated by the transformer-based sensor fusion~\cite{prakash2021multi}, we implement a variant of LI-Fusion and IL-Fusion modules by replacing the original attention mechanism with cross-attention. 
However, we notice that cross-attention even leads to slightly worse performance. We assume the reason is that cross-attention neglects the valuable preliminary information of one-to-one correspondence provided by a projection matrix.}

\noindent\zliu{\textbf{Voxel-based CB-Fusion.} We also investigate the application of CB-Fusion in the popular voxel-based frameworks~\cite{yan2018second,deng2021voxel,yin2021center}. Concretely, We insert the CB-Fusion module after the first 3D convolution layer of the 3D backbone, since the high-resolution voxel feature is more valuable for reducing the misalignment between two modalities. We select the one-stage detector of SECOND~\cite{yan2018second} and the two-stage detector of Voxel-RCNN~\cite{deng2021voxel} as LiDAR-only baseline. As shown in Table~\ref{voxel_kitti}, CB-Fusion brings noticeable improvement on Cars for all difficulty levels on the KITTI dataset, which illustrates the effectiveness of CB-Fusion.
\zliunew{Besides, we provide the results when the GT sampling strategy~\cite{yan2018second} is employed for reference inspired by MoCa~\cite{zhang2020multi}. Under this setting, CB-Fusion also leads to consistent improvements and boosts the performance of Voxel-RCNN from 84.49\% to 85.20\% on moderate Cars.}
}

\begin{table}[htbp] 
\scriptsize
\centering
\caption{Results on the Waymo validation dataset with a single frame of 20\% training samples. We use the official mAP/mAPH~\cite{sun2020scalability} as the evaluation metric.  Vec-(L1/L2) and Ped-(L1/L2) means the Vehicle and Pedestrians on the difficulty of LEVEL1/LEVEL2. \zliunew{$\dag$ means using a deeper feature extractor~(Mask-RCNN~\cite{he2017mask} based on ResNet-50~\cite{he2016deep}) in the image stream.} }
\scalebox{0.95}{\begin{tabular}{c|c|c|c|c}
\hline
Methods & Vec-L1 & Vec-L2 & Ped-L1 & Ped-L2   \\
\hline
CenterPoint~\cite{yin2021center}   &71.29/70.72  &63.19/62.68  & 71.61/64.75 &63.70/57.45 \\  
+ CB-Fusion  &\textbf{71.67/71.09}  &\textbf{63.61/63.08}  & \textbf{71.97/65.35}   &\textbf{63.97/57.93}  \\
+ CB-Fusion$^\dag$ &\zliunew{\textbf{71.84/71.31}}  & \zliunew{\textbf{63.82/63.33}} & \zliunew{\textbf{72.68/65.99}} &\zliunew{\textbf{64.83/58.70}}  \\
\hline
Voxel RCNN~\cite{deng2021voxel}   &76.14/75.67  &67.74/67.31  & 78.31/71.22   &69.72/63.19  \\
+ CB-Fusion  &\textbf{76.33/75.85}  &\textbf{67.91/67.48}  & \textbf{78.75/72.09}   &\textbf{70.17/64.02}  \\
+ CB-Fusion$^\dag$  &\zliunew{\textbf{76.57/76.10}}  &\zliunew{\textbf{68.29/67.86}}  & \zliunew{\textbf{79.30/72.83}}  &\zliunew{\textbf{70.78/64.77}}  \\
\hline
\end{tabular}}
\label{voxel_waymo}
\end{table}

\noindent\zliu{\textbf{CB-Fusion on Large-scale Dataset.} \zliunew{To further illustrate the scalability of our method, we apply our CB-Fusion module to the challenging large-scale Waymo dataset~\cite{sun2020scalability}. As shown in Table~\ref{voxel_waymo}, CB-Fusion leads to consistent improvements across different backbones, categories, and difficulty levels. Moreover, we notice that a deeper 2D feature extractor is important for better performance. As shown in Table~\ref{voxel_waymo}, replacing the default tiny image backbone with Mask-RCNN~\cite{he2017mask} based on ResNet-50~\cite{he2016deep}, CB-Fusion leads to noticeable improvements of 0.55\% and 1.58\% over Voxel-RCNN on Vehicles and Pedestrians in terms of L2 mAPH. In this paper, we mainly focus on the design of the fusion mechanism and leave the usage of a large image backbone as future work.}}

\subsection{Analysis of Visualization} \label{exp_vis}


\begin{figure*}[htbp]
\centering
  \includegraphics[width=0.96\linewidth]{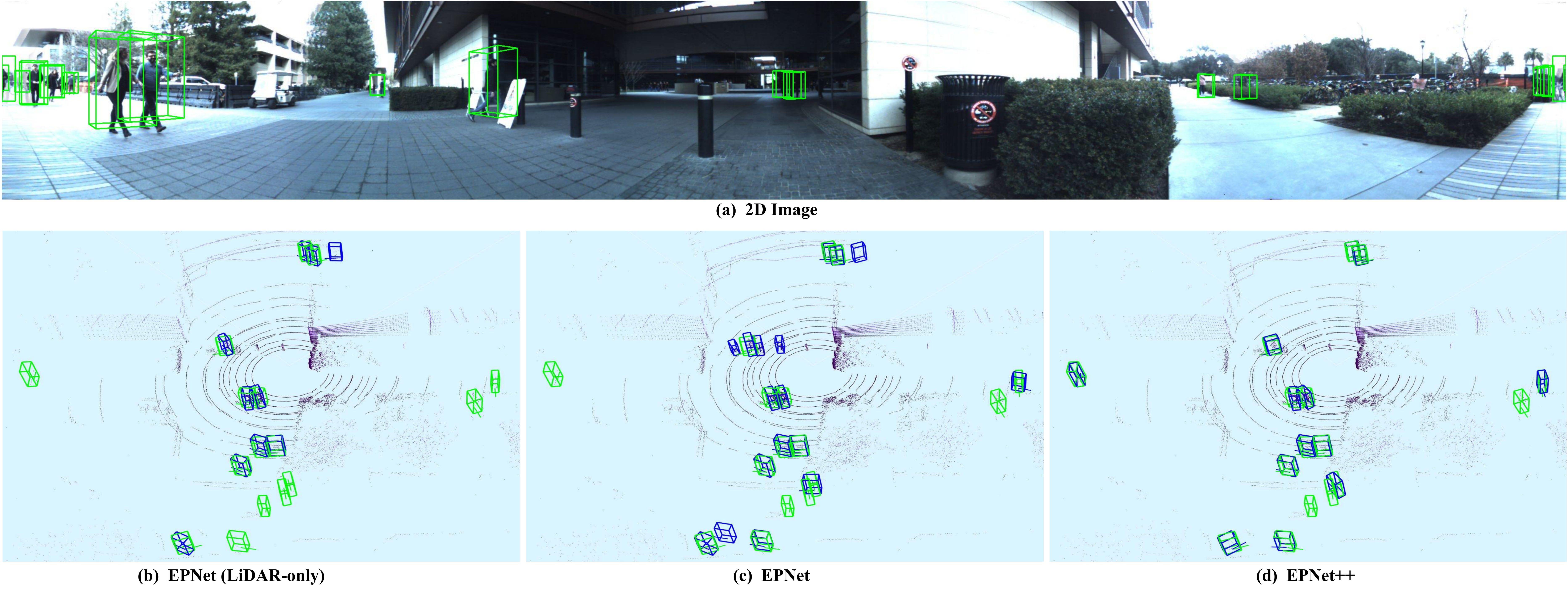}
\caption{The visualization results on the JRDB dataset. The predicted boxes and GT boxes are highlighted in blue and green, respectively. In the first row, we project the GT 3D bounding boxes onto the $360^{\circ}$ cylindrical image. The second row shows the GT boxes and the prediction results of EPNet~(LiDAR-only), EPNet and the proposed EPNet++, respectively.}
\label{via_jrdb_results}
\end{figure*}

\begin{figure}[ht]
\centering
  \includegraphics[width=0.96\linewidth]{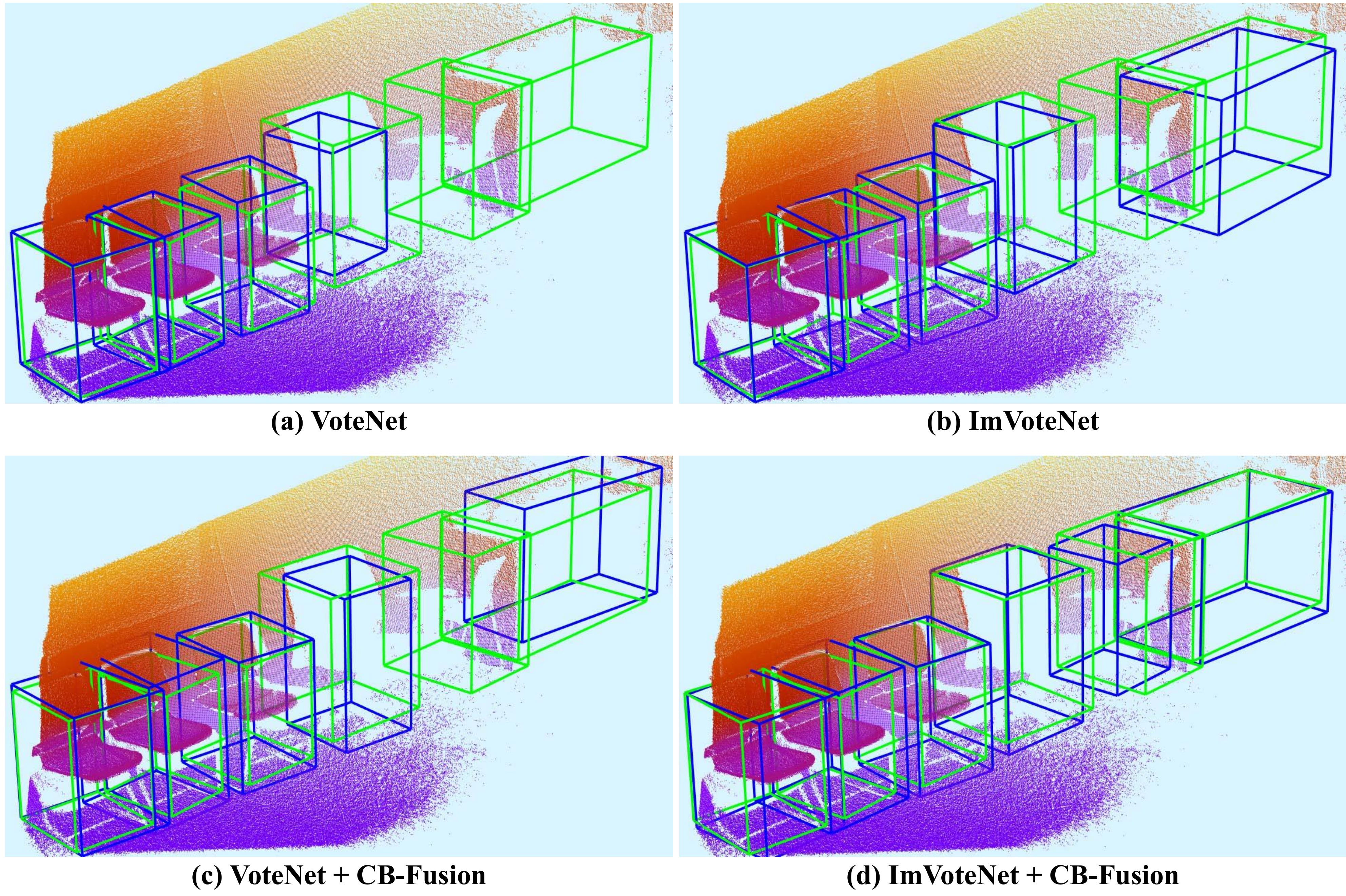}
\caption{The visualization results on the SUN-RGBD dataset. The first row displays the 3D detection boxes of VoteNet~\cite{qi2019deep} and ImVoteNet~\cite{qi2020imvotenet}. The second row presents the results of integrating our CB-Fusion module into these two approaches.
The predicted boxes and GT boxes are highlighted in blue and green, respectively.}
\label{via_sungrbd_results}
\end{figure}

\begin{figure}[htbp]
\begin{center}
  \includegraphics[width=0.96\linewidth]{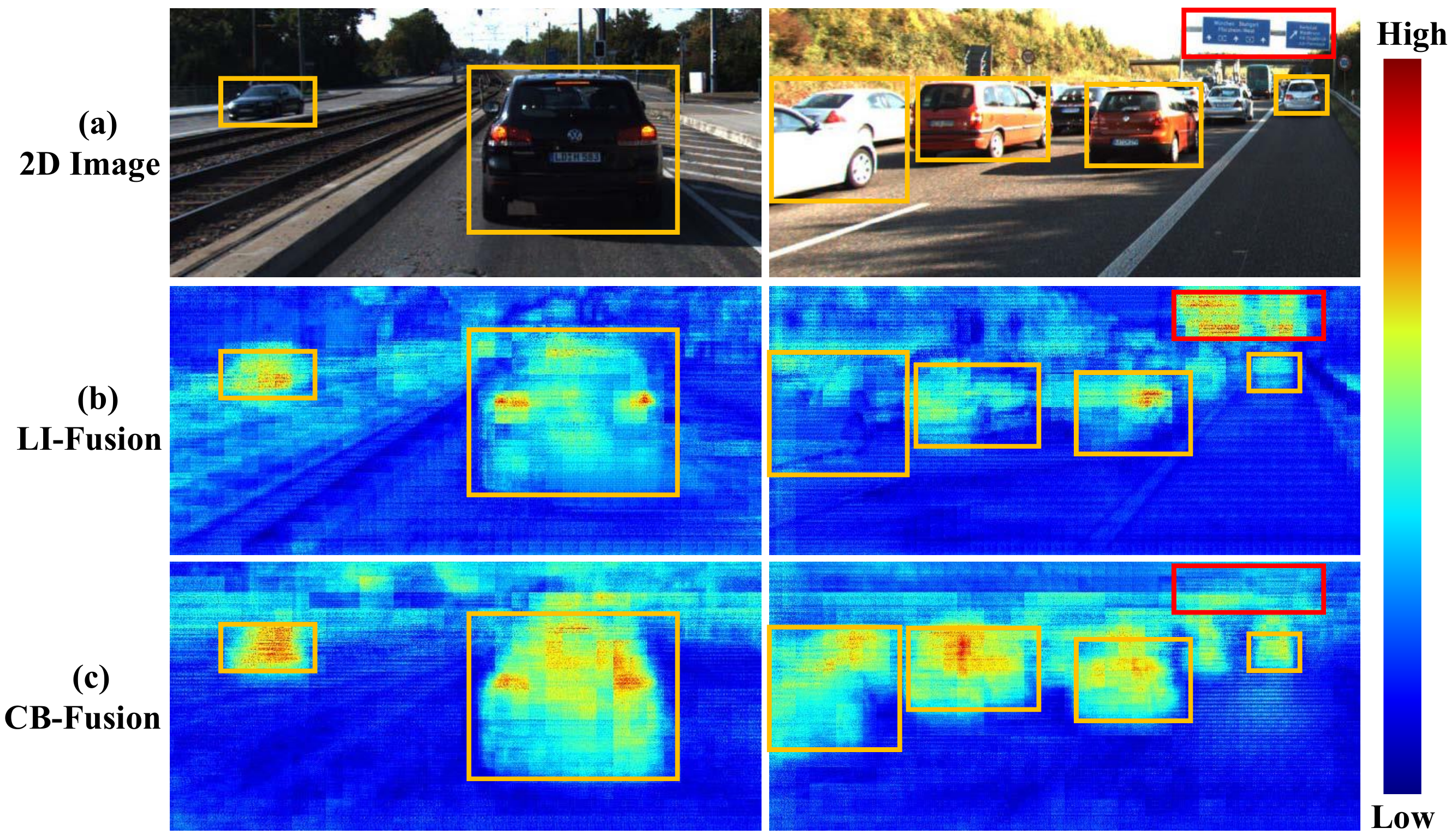}
\end{center}
\caption{Visualization of the learned semantic image feature. Foreground objects are highlighted with yellow rectangle boxes. The red rectangle box marks the bad region learned from the image stream.}
\label{vis_kitti_feat_results}
\end{figure}

\subsubsection{Visualization of Detection Results}

\noindent\textbf{KITTI Dataset.}~Fig.~\ref{vis_kitti_results} presents qualitative comparison for LiDAR-only EPNet~(removing the image stream), EPNet, and EPNet++ on the KITTI dataset. As is shown in the first column, LiDAR-only EPNet fails to detect the pedestrians far away with few points. With the aid of multi-modal fusion, EPNet and EPNet++ detect these pedestrians by leveraging the semantic information from images. In the second column, our EPNet++ even detects a 3D object that is labeled as \textit{DontCare}~\cite{geiger2012we}\footnote{KITTI Dataset uses ``DontCare" to denote regions in which objects have not been labeled,
which are usually too far away from the laser scanner.} in GT label. In the last column, our methods produce a failure case, where a long-distance pedestrian is not recognized by our detector. 
The possible reason is that this distant pedestrian with only occupied few points and maybe be interfered by its nearby objects.

\noindent\textbf{JRDB Dataset.}~As is shown in Fig.~\ref{via_jrdb_results}, EPNet detects more pedestrians than LiDAR-only EPNet with the help of the LI-Fusion module. However, EPNet also produces several false positives for challenging objects with occlusion, and/or few points. In contrast, EPNet++ successfully filters these false positives thanks to both the more discriminative feature representations generated by superior CB-Fusion and more reliable confidence scores optimized by MC loss.

\noindent\textbf{SUN-RGBD Dataset.}~The Fig.~\ref{via_sungrbd_results} illustrates the qualitative results of two representative detectors VoteNet~\cite{qi2019deep}, ImVoteNet~\cite{qi2020imvotenet} under two settings, \textit{i.e.}, with and without our CB-Fusion module. It can be observed that CB-Fusion leads to more precise boxes and higher recall, further demonstrating the effectiveness and generalization of our CB-Fusion module.




\subsubsection{Visualization of Learned Image Semantic Features}

To investigate the image semantic features learned by the LI-Fusion and CB-Fusion modules, we remove the explicit supervision information~(\textit{i.e.}, 2D semantic segmentation annotations) from the two-stram RPN. It means that the image stream is optimized with only the implicit supervision from the geometric stream of the two-stream RPN. 
\zliu{We present the visualization of the learned image feature map by summing all the channel maps, as shown in Fig.~\ref{vis_kitti_feat_results}.}
Although no explicit supervision is applied, surprisingly, the image stream learns well to differentiate the foreground objects and extracts rich semantic features from camera images. It demonstrates that the proposed LI-Fusion and CB-Fusion modules accurately establish the correspondence between LiDAR point cloud and camera image, thus can provide the complement semantic image information to enhance the point features. Besides, in the second column, compared with the LI-Fusion module, the learned semantic feature through the CB-Fusion module is more plentiful and can effectively suppress the bad region.

\section{Conclusion}
This paper has presented a novel end-to-end trainable framework for multi-modal 3D object detection named EPNet++, consisting of a two-stream RPN and a refinement network. Specifically, we propose a novel CB-Fusion module to enable bi-directional feature enhancement pathways, leading to more discriminative and comprehensive feature representations. The MC loss is utilized to promote the consistency of the confidence scores between the image and geometric streams, which is beneficial for selecting high-quality proposals.
Extensive ablation studies are conducted to validate the effectiveness of the proposed CB-Fusion module and the MC loss. Our method achieves competitive and even SOTA detection performance on the KITTI, JRDB, and SUN-RGBD datasets, which demonstrates the superiority of EPNet++. Moreover, EPNet++ outperforms the SOTA methods with remarkable margins in highly sparse point cloud scenes, benefiting from the powerful CB-Fusion module and MC loss.
In the future, we would like to investigate more efficient architecture for multi-modal fusion through unifying different sensors into one stream.

\section{Acknowledgement}
This work was supported by the National Science Foundation of China for Distinguished Young Scholars (No. 62225603).

{\small
	\bibliographystyle{IEEEtran}
	\bibliography{ref}
}

\end{document}